%% file: main.tex
\documentclass{article}
\usepackage[preprint]{neurips_2022}
\usepackage[utf8]{inputenc} 
\usepackage[T1]{fontenc}    
\usepackage{hyperref}       
\usepackage{url}            
\usepackage{booktabs}       
\usepackage{amsfonts}       
\usepackage{nicefrac}       
\usepackage{microtype}      
\usepackage{xcolor}         
\usepackage{multirow}

\usepackage{todonotes}
\usepackage[normalem]{ulem}

\usepackage[acronym]{glossaries}
\usepackage{amsmath,amsthm}
\usepackage{amssymb}
\usepackage[ruled,vlined]{algorithm2e}
\usepackage{subfig}

\setcitestyle{numbers}

\makeglossaries
\newacronym{SSL}{SSL}{Self-Supervised Learning}

\title{Benchmark for Uncertainty \& Robustness in \\ Self-Supervised Learning}

\author{%
	Ha Manh Bui$^{1}$ \hspace{0.1cm}
	Iliana Maifeld-Carucci$^{1}$ \hspace{0.1cm}\\
	$^{1}$ Department of Computer Science, Johns Hopkins University, Baltimore, Maryland, USA\\
	\texttt{\{hb.buimanhha@gmail.com, imaifel1@jhu.edu\}}
}
\begin{document}

\maketitle

\input{paragraphs/0_abstracts.tex}
\input{paragraphs/1_introduction.tex}
\input{paragraphs/2_related_work.tex}

\input{paragraphs/3_method.tex}
\input{paragraphs/4_experiments.tex}
\input{paragraphs/5_conclusion.tex}
\newpage
\bibliographystyle{unsrt}
\bibliography{refs}

\input{appendix/app_main.tex}

\end{document}

%% file: paragraphs/0_abstracts.tex
\begin{abstract}
\acrfull{SSL} is crucial for real-world applications, especially in data-hungry domains such as healthcare and self-driving cars. In addition to a lack of labeled data, these applications also suffer from distributional shifts. Therefore, an \acrshort{SSL} method should provide robust generalization and uncertainty estimation in the test dataset to be considered a reliable model in such high-stakes domains. However, existing approaches often focus on generalization, without evaluating the model’s uncertainty. The ability to compare \acrshort{SSL} techniques for improving these estimates is therefore critical for research on the reliability of self-supervision models. In this paper, we explore variants of \acrshort{SSL} methods, including Jigsaw Puzzles, Context, Rotation, Geometric Transformations Prediction for vision, as well as BERT and GPT for language tasks. We train \acrshort{SSL} in auxiliary learning for vision and pre-training for language model, then evaluate the generalization (in-out classification accuracy) and uncertainty (expected calibration error) across different distribution covariate shift datasets, including MNIST-C, CIFAR-10-C, CIFAR-10.1, and MNLI. Our goal is to create a benchmark with outputs from experiments, providing a starting point for new \acrshort{SSL} methods in Reliable Machine Learning. All source code to reproduce results is available at \href{https://github.com/hamanhbui/reliable_ssl_baselines}{https://github.com/hamanhbui/reliable\_ssl\_baselines}.
\end{abstract}

%% file: paragraphs/1_introduction.tex
\section{Introduction}
\acrfull{SSL} has recently become an important research direction in machine learning due to no label requirements and being close to how humans learn and generalize. In \acrshort{SSL}, the model is optimized with sample and new task self-supervised labels, with the goal of the model learning meaningful representations from unlabeled data. Without label requirements, \acrshort{SSL} becomes a potential direction to create a generalized model for a number of real-world applications such as cancer diagnostic imaging in health care, driver monitoring systems in self-driving cars, weather forecasts in meteorology, etc. Many of these applications, however, not only often lack labeled data but also suffer from distributional shifts by observed or unobserved confounding variables. For instance, in cancer diagnostic imaging, different patients have different physical conditions, and different hospitals have different camera types, leading to covariate-shift problems.

In order to have a reliable model in these mentioned high-stake applications, \acrshort{SSL} needs to satisfy both conditions of robust generalization and uncertainty estimation under distributional shifts~\cite{plex}. Specifically, robust generalization means the model can avoid over-fitting and generalize well under distributional shifts. Meanwhile, uncertainty estimation means the model can avoid over/under-confident predictions and provide predictive uncertainty, i.e., showing how the confidence level about the prediction, deferring to human experts when necessary. Regarding generalization under the distributional shift, prior works have shown the advantages of \acrshort{SSL} to boost generalization in many covariate-shift settings, especially by using \acrshort{SSL} as an auxiliary task to jointly train with the main task~\cite{carlucci2019domain, NEURIPS2019_a2b15837} for vision and pre-training then fine-tuning for language tasks~\cite{devlin-etal-2019-bert,lewis-etal-2020-bart,NEURIPS2020_1457c0d6}. Regarding combining with uncertainty estimation, the work of~\cite{NEURIPS2019_a2b15837} recently shows the potential of using auxiliary training with \acrshort{SSL} to improve model robustness and uncertainty. However, they only measure the performance in terms of out-of-distribution detection, without measuring the calibration ability. Moreover, they only consider the vision task and evaluate the generalization in the unrealistic scenario with independent and identically distributed test datasets. Meanwhile, for \acrshort{SSL} in language tasks, it is only quite recently that research has comprehensively investigated the calibration of the pre-trained language model pipeline \cite{xiao2022uncertainty}.

Motivated by the idea of using \acrshort{SSL} to create a reliable model, in this paper, we propose bench-marking for \acrshort{SSL} methods, including Jigsaw Puzzles, Context, Rotation, and Geometric Transformations Prediction for vision, as well as BERT/GPT for language tasks. For vision, we train \acrshort{SSL} as an auxiliary task to jointly train with the main task. For language, we do pre-training and then fine-tuning for the downstream task of natural language inference. We then compare these methods based on their generalization capabilities with in-out classification accuracy and uncertainty with expected-calibration error across covariate shift datasets, including corrupted MNIST, corrupted CIFAR-10 with five skew intensities, real-world shift in CIFAR-10.1, and real-world shift in MNLI. Our goal includes first creating a benchmark for a fair comparison between \acrshort{SSL} methods in reliable machine learning, including source code, model checkpoints, experiment outputs, and comparing generalization and uncertainty results. Secondly, we find the answers to the following questions: \textit{"Does \acrshort{SSL} always improve generalization?"}, \textit{"Can \acrshort{SSL} improve predictive uncertainty?"}, and \textit{"What is the best reliable machine learning method in \acrshort{SSL}?"}.

%% file: paragraphs/2_related_work.tex
\section{Background and Related work}
\subsection{Reliable Machine Learning}
Reliable machine learning has recently been defined by two necessary conditions: uncertainty and robust generalization~\cite{abraham1983statistical, Ghahramani:2015:Nature:26017444,tran-2020-tutorial,plex}. Robust generalization involves an estimate or forecast about an unseen event~\cite{Ghahramani:2015:Nature:26017444}. Diverse methods try to achieve this property by creating algorithms in different settings related to generalization~\cite{domainbed,NEURIPS2021_b0f2ad44}. Meanwhile, uncertainty involves imperfect or unknown information where it is impossible to exactly describe an existing state~\cite{abraham1983statistical,tran-2020-tutorial}. It includes two types of uncertainty: aleatoric uncertainty, and epistemic uncertainty. Aleatoric uncertainty which is also called data uncertainty means uncertainty from the natural stochasticity of observations and is often irreducible with more data~\cite{DBLP:conf/uai/SmithG18}. For example, the random observations of fair flipping coins. In contrast, epistemic uncertainty a.k.a., model uncertainty is uncertainty from a lack of observations, and is reducible with more data, such as outlier data~\cite{DBLP:conf/uai/SmithG18}. These uncertainties are implied in predictive uncertainty, however, are often non-trivial to disentangle from the softmax entropy of neural networks in general~\cite{kirsch2021pitfalls,DBLP:conf/uai/SmithG18}.

In this paper, we measure the generalization by in-out classification accuracy, which means accuracies in both in-data distribution and out-of-data distribution. A reliable model is expected to achieve high accuracy in both of these two distributions of data to satisfy the robust generalization condition~\cite{NEURIPS2021_b0f2ad44}. Meanwhile, we measure the Expected Calibration Error (ECE)~\cite{ECE} to reflect the model uncertainty by using the deterministic method with the max of predictive softmax~\cite{ECENN}. A model is considered reliable if achieves low ECE in both in-data and out-of-data distribution sets.

\subsection{Benchmark for Uncertainty \& Robustness}
With the crucial nature of reliable models in high-stakes real-world applications, requirements for the ability to compare reliable deep learning techniques for improving uncertainty and robustness have been considered in~\cite{nado2021uncertainty,plex}. In particular, the work of~\cite{nado2021uncertainty} created a high-quality implementation of the standard deterministic and probabilistic methods on out-of-distribution tasks for text and image classification. Specifically, they include predictive metrics such as accuracy and uncertainty metrics with selective prediction and calibration error, compute metrics with inference latency, and evaluate performance under in- and out-of-distribution MNIST, CIFAR, and ImageNet datasets. Later on, the paper~\cite{plex} provided more comparisons of retrained large model extensions. The value of~\cite{nado2021uncertainty,plex} is that they not only compare performance across benchmarks, but also provide high-quality source code, a fair comparison across different settings, and reliable results to support the research community in having a clear picture about where reliable neural networks is going. 

Similarly, with the same motivation of~\cite{nado2021uncertainty,plex}, we provide our source code, fair implementation settings, and output results. However, differently, we do experiments on a different type of \acrshort{SSL} and provide a Pytorch implementation while~\cite{nado2021uncertainty, plex} compares across probabilistic deep supervised learning and publish a Tensorflow implementation. Our goal additionally discovers whether \acrshort{SSL} really helps in both generalization and uncertainty estimation, which has been discovered by recent work~\cite{NEURIPS2019_a2b15837, xiao2022uncertainty}, and if yes, figure out which \acrshort{SSL} method is the best for a reliable machine learning system.

\subsection{Benchmark for Self-Supervised Learning}
Benchmark for general \acrshort{SSL} has been investigated in~\cite{BENCHMARKS2021_b3e3e393, agnostic-bm, scalingbm, xiao2022uncertainty}. However, these works often consider the pre-trained \acrshort{SSL} model in transfer learning, and often for language and speech domain fine-tuning. For instance,~\cite{BENCHMARKS2021_b3e3e393} do benchmarking in pre-trained \acrshort{SSL} wav2vec models, in four downstream tasks along with automatic speech recognition, spoken language understanding, automatic speech translation, and automatic motion recognition. Image tasks are considered in~\cite{agnostic-bm} for domain-agnostic benchmarking for \acrshort{SSL}, however, in multi-modality settings with other tasks like multichannel sensor data, English text, speech recordings, multilingual text, chest x-rays, and images with text descriptions. Moreover, this work only compares across contrastive learning algorithms and transformer architectures, ignoring hand-crafted pretext tasks for vision, and also the need to use the pre-trained features for downstream tasks. Most similar to our vision benchmarking is~\cite{scalingbm}, which compares Jigsaw Puzzle and Colorization in hand-crafted pretext tasks \acrshort{SSL} and treats \acrshort{SSL} as a pre-training task and does fine-tuning for a downstream task later.

Our paper is different from these mentioned works in terms of the vision task, we are motivated by jointly training \acrshort{SSL} with the main task, treating \acrshort{SSL} as an auxiliary task to boost generalization performance~\cite{carlucci2019domain, NEURIPS2019_a2b15837}. This is more novel because we do not need to do additional fine-tuning steps, and is comparable with the Empirical Risk Minimization setting~\cite{abraham1983statistical}. More importantly, we do extensive experiments in more hand-crafted pretext tasks \acrshort{SSL} techniques, including Jigsaw Puzzle, Context, Rotation, and Geometric Transformations Prediction. Our goal is to find what is the best techniques for the auxiliary task, reducing computational cost and training time in the vision task. For the language task, our paper aims to verify the findings of \cite{xiao2022uncertainty}, while looking at the generalization and calibration abilities of GPT2.

%% file: paragraphs/3_method.tex
\section{Method}
In this section, we will formalize the problems of distributional shifts, then introduce our training procedure when combined with \acrshort{SSL} for each vision and language task, and finally summarize each instance algorithm in \acrshort{SSL} families.

\subsection{Distributional Shift}
\textbf{Notation.} Let $\mathcal{X} \subset \mathbb R^D$ be the sample space and $\mathcal{Y} \subset \mathbb R$ be the label space. Denote the set of joint probability distributions on $\mathcal{X} \times \mathcal{Y}$ by $\mathcal{P}_{\mathcal{X} \times \mathcal{Y}}$, and the set of probability marginal distributions on $\mathcal{X}$ by $\mathcal{P}_{\mathcal{X}}$. A dataset is defined by a joint distribution $p(x, y) \in \mathcal{P}_{\mathcal{X}\times \mathcal{Y}}$, and let $\mathcal P$ be a measure on $\mathcal{P}_{\mathcal{X} \times \mathcal{Y}}$, i.e., whose realizations are distributions on $\mathcal{X} \times \mathcal{Y}$. Denote training data by $D_{s} = \{(x_{i}, y_{i})\}_{i=1}^{N}$, where $N$ is the number of data points in $D_{s}$, i.e., $(x_{i}, y_{i}) \overset{iid}{\sim} p_s(x,y)$ where $p_s(x,y) \sim \mathcal P$.

\textbf{Problem definition.}
In a typical out of distribution (O.O.D.) framework, a learning model which is only trained on the training data $D_{s}$, arrives at a good generalization performance on the test dataset $D_{t} = \{(x_{i}^t, y_{i}^t)\}_{i=1}^{N_T}$, where $(x_{i}^t, y_{i}^t) \overset{iid}{\sim} p_t(x,y)$ and $p_t(x,y) \sim \mathcal{P}$. In this paper, we experiment with covariate shift where the distribution of inputs changes: $p_t(x) \neq p_s(x)$, while the conditional distribution of outputs is unchanged: $p_t(y\mid x) = p_s(y\mid x)$.\\
For example, in the CIFAR-10.1 dataset, the training set includes natural horse images and the new input is a drawing of that horse. In order to be robust under covariate shift, a model should be able to reliably make correct predictions on noisy, corrupted, and otherwise distribution-shifted inputs.

\subsection{Auxiliary Self-Supervised Training for Vision}
\begin{figure}[ht!]
\vskip -0.1in
\begin{center}
\includegraphics[width=1.0\linewidth]{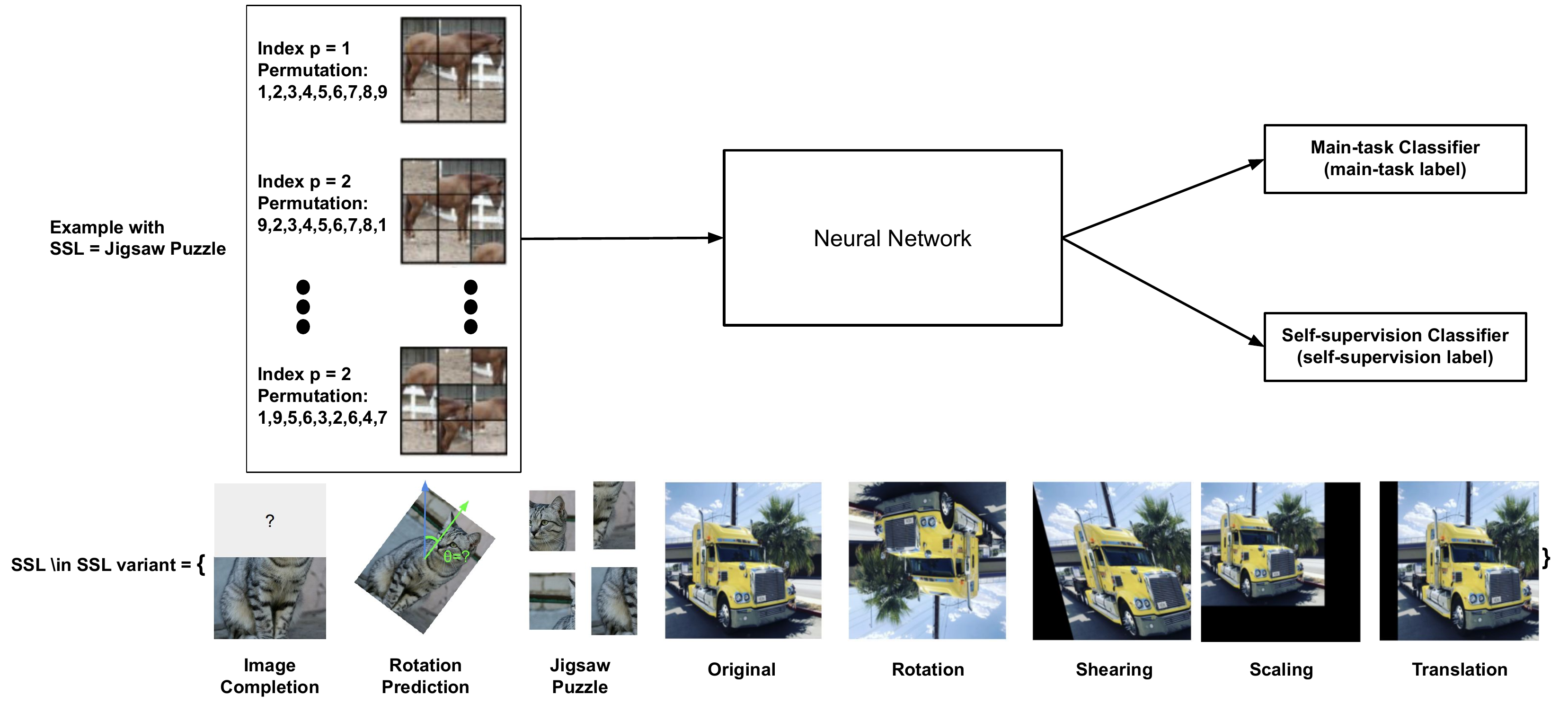}
\end{center}
   \caption{The overview of our training procedure in the vision task where we jointly optimize the objective function of the classifier with main task labels and the self-supervision classifier with self-supervision labels. We explore four different hand-craft~\acrshort{SSL} methods for the vision task, including Context Prediction, Rotation Prediction, Geometric Transformations Prediction, and Jigsaw Puzzles. The above architecture is an example of the Jigsaw Puzzles method.}
\label{fig:teaser}
\vskip -0.2in
\end{figure}
We train Self-supervision as an auxiliary task in multi-task learning. We summarize our training method in Figure~\ref{fig:teaser}. In particular, we jointly optimize the objective function of the classifier with main task labels and the self-supervision classifier with self-supervision labels. Formally, given a model which composites a feature extractor $f: \mathcal{X} \rightarrow \mathcal{Z}$, where $\mathcal{Z}$ is the latent space, classifier embedding $h: \mathcal{Z} \rightarrow \mathcal{Y}$, and self-supervision classifier $g: \mathcal{Z} \rightarrow \mathcal{S}$, where $\mathcal{S}$ is the self-supervision label space, we optimize the following objective function:
\begin{equation}
\label{objective_function}
\underset{\theta_{f,h,g}}{\operatorname{min}} \mathbb{E}_{(x,y)\in D_{B}} \left [ -y \log \left ( h(f(x)) \right ) \right ] +
\mathbb{E}_{(x,s)\in D_{B}} \left [ -s \log \left ( g(f(t(x))) \right ) \right ],
\end{equation}
where $t$ is a transformation of a self-supervised signal. Depending on each \acrshort{SSL} method, the function $t$ and space $\mathcal{S}$ are different. We next informally describe each of these~\acrshort{SSL} methods in the following section.

\textbf{Self-Supervised Methods for Vision}
Figure~\ref{fig:teaser} summarizes 4 different variants of \acrshort{SSL} we consider, including:
\begin{itemize}
    \item \textbf{Context Prediction~\cite{contextprediction}.}
In context prediction, the self-supervision task for learning patch representations involves randomly sampling a patch and then one of eight possible neighbors of the original image. The function $t$ transforms the original image $x$ to split patches, and the supervision label $s$ is the label that represents the position of the patch. In our experiments, the transformation $t$ transforms $x$ into 9 split patches. 
    \item \textbf{Rotation Prediction~\cite{TransformNet}.}
The rotation prediction considers learning to predict the rotated self-supervision from the original input. In particular, the function $t$ transforms the original image $x$ by rotating with a specific degree, and the supervision label $s$ is the label that represents the degree of the rotation. In our implementation, we consider 8 degrees, including $\left \{ 0, 45, 90, 135, 180, 225, 270, 315 \right \}$.
    \item \textbf{Geometric Transformations Prediction~\cite{TransformNet}.}
The task of Geometric Transformations is to predict what transformation techniques were applied to the original image. The transformation algorithms applied for $t$ include rotation, shearing, scaling, and translation. This is therefore a multi-label classification problem where the label $s$ can be more than one for this transformation.
    \item \textbf{Jigsaw Puzzles~\cite{carlucci2019domain}.}
Transformation $t$ in Jigsaw Puzzles splits original images into different patches, then shuffle these patches. The label $s$ is then generated by using the maximal
Hamming distance algorithm in~\cite{Unsupervised-Jigsaw}, with a set of P patch permutations and assigns an index to each of them. In our experiment, we consider 9 patches and 30 permutations, casting it into a classification problem.
\end{itemize}

\subsection{Pre-training Self-Supervised and Fine-Tuning for Language}
Similar to \cite{plex}, for SSL in language, we investigate the real-world task of natural language inference. This is a multi-class classification task mapping a premise and hypothesis to one of the categories of entailment, contradiction, or neutral. Language models differ greatly in their structure and objectives and for this work, we wanted to consider two different models. The first, BERT, short for Bidirectional Encoder Representations from Transformers, is an encoder only, masked language model while the second, GPT2, is a decoder only language modeling model. For both models we use the pre-trained model with a linear classification layer on top to fine-tune the models for natural language inference.

%% file: paragraphs/4_experiments.tex
\section{Experiments}\label{sec:experiments}
\subsection{Experimental settings}
\textbf{Datasets.} To compare the effectiveness of \acrshort{SSL} methods under distributional shifts for vision, we utilize 3 standard commonly used datasets, including: \textbf{MNIST-C}~\cite{mu2019mnist} contains 10,000 samples of dimension $(1, 28, 28)$ and 10 classes, shifted from 15 corruptions to the test images in the MNIST dataset~\cite{mnist}, \textbf{CIFAR-10-C}~\cite{hendrycks2018benchmarking} includes 10,000 images of dimension $(3, 32, 32)$ and 10 classes, generated from 15 common corruptions, however adding 4 more extra corruptions to the test images in the CIFAR-10 dataset~\cite{cifar10}, \textbf{CIFAR-10.1}~\cite{recht2018cifar10.1} contains two versions: v4 includes 2,021 images, v6 contains 2,000 images with exactly class balanced, both of dimension $(3, 32, 32)$ and 10 classes, are new real-world images that are a subset of the TinyImages dataset~\cite{torralba2008tinyimages}, designed to minimize distribution shift relative to the original dataset. For the \acrshort{SSL} language task, the widely used \textbf{MultiNLI} dataset includes 433K sentence pairs and 3 classes~\cite{williams2017broad}. The detail of each dataset is provided in Appendix~\ref{apd:dataset}.

\textbf{Implementation.} In the vision task, we use the cross-validation technique for model selection. In particular, for all datasets, we first merge the raw training and validation, then, we run the test 10 times with 10 different seeds. For each random seed, we randomly split training and validation and choose the model maximizing the accuracy on the validation set, then compute performance on the given test sets. The mean and standard deviation of classification accuracy from these 10 runs are reported. We evaluate performance based on backbones LetNet5~\cite{lenet5} for MNIST datasets and Wide Resnet-28-10~\cite{wideresnet} for CIFAR10 datasets to compare the methods. For the MultiNLI dataset, the training set is composed of written language, while the text set is transposed verbal language, representing a mild, naturally-occurring covariate shift. Data-processing techniques, model architectures, and hyper-parameters are presented in detail in Appendix~\ref{apd:implementation}. All source code to reproduce results are available at~\href{https://github.com/hamanhbui/reliable_ssl_baselines}{https://github.com/hamanhbui/reliable\_ssl\_baselines}, all checkpoints are available at~\href{https://drive.google.com/drive/folders/1Goy76FCp8NmlQiwAuQdj7HETOy85jLMa?usp=share_link}{this google drive}.
\subsection{Results}
\subsubsection{Vision Task}
\begin{table}[ht!]
\caption{Results for LeNet5 on MNIST and  Wide Resnet 28-10 on CIFAR-10, averaged over 10 seeds: negative log-likelihood (lower is better), accuracy (higher is better), and expected calibration error (lower is better). cNLL, cAccuracy, and cECE are NLL, accuracy, and ECE averaged over MNIST-C’s and CIFAR-10-C’s
corruption types, and CIFAR-10.1 v4 and v6 version.}
\label{tab:results}
\centering
\begin{tabular}{lcccccc}
\toprule
\textbf{Method} & \textbf{NLL($\downarrow$)} & \textbf{Accuracy($\uparrow$)} & \textbf{ECE($\downarrow$)} & \textbf{cNLL($\downarrow$)} & \textbf{cAccuracy($\uparrow$)} & \textbf{cECE($\downarrow$)}\\
\midrule
\multicolumn{7}{c}{MNIST \& MNIST-C}\\
\midrule
ERM~\cite{vapnik1998erm} & 0.0365 & 98.87 & 0.0107 & 1.5439 & \textbf{77.64} & \textbf{0.1759}\\
Context~\cite{contextprediction} & 0.0326 & \textbf{99.02} & \textbf{0.0099} & 2.1087 & 71.76 & 0.2091\\
Rotation~\cite{TransformNet} & 0.0373 & 98.83 & 0.0119 & 1.9216 & 69.28 & 0.2085\\
Affine~\cite{TransformNet} & \textbf{0.0306} & 99.01 & 0.0105 & \textbf{1.5241} & 72.33 & 0.1792\\
Jigsaw~\cite{carlucci2019domain} & 0.0372 & 98.86 & 0.0114 & 1.9792 & 71.36 & 0.2042\\
\midrule
\multicolumn{7}{c}{CIFAR-10 \& CIFAR-10-C}\\
\midrule
ERM~\cite{vapnik1998erm} & 0.9988 & 86.85 & 0.1115 & 3.4181 & 66.22 & 0.2858\\
Context~\cite{contextprediction} & 3.0424 & 72.17 & 0.2414 & 5.1379 & 58.06 & 0.3663\\
Rotation~\cite{TransformNet} & \textbf{0.5633} & 89.33 & \textbf{0.0820} & 2.6236 & 64.40 & 0.2725\\
Affine~\cite{TransformNet} & 0.7394 & 89.10 & 0.0908 & 2.7986 & 68.71 & 0.2580\\
Jigsaw~\cite{carlucci2019domain} & 0.7316 & \textbf{90.12} & 0.0851 & \textbf{2.5242} & \textbf{71.02} & \textbf{0.2407}\\
\midrule
\multicolumn{7}{c}{CIFAR-10 \& CIFAR-10.1}\\
\midrule
ERM~\cite{vapnik1998erm} & 0.9988 & 86.85 & 0.1115 & 1.9978 & 76.05 & 0.1993\\
Context~\cite{contextprediction} & 3.0424 & 72.17 & 0.2414 & 4.5094 & 61.17 & 0.3372\\
Rotation~\cite{TransformNet} & \textbf{0.5633} & 89.33 & \textbf{0.0820} & \textbf{1.2327} & 78.92 & \textbf{0.1584}\\
Affine~\cite{TransformNet} & 0.7394 & 89.10 & 0.0908 & 1.5512 & 79.25 & 0.1706\\
Jigsaw~\cite{carlucci2019domain} & 0.7316 & \textbf{90.12} & 0.0851 & 1.5763 & \textbf{80.97} & 0.1628\\
\bottomrule
\end{tabular}
\end{table}

Table~\ref{tab:results} summarizes the results of our experiments on 3 benchmark datasets when compared between these~\acrshort{SSL} models in 3 criteria, including Negative Log Likelihood (NLL), Accuracy, and Expected Calibration Error (ECE)~\cite{ECE}. The full result per dataset are in~\href{https://docs.google.com/spreadsheets/d/1QdYtDm_GA76hfpso8RoAeysn3Yg3WPeJ91VXahhoPgM/edit?usp=share_link}{this google excel} and the box-plots statistical comparison per each shift-intensity of CIFAR-10-C is provided in Figure~\ref{fig:dcifar-skew} in Appendix~\ref{apd:results}. From these results, we draw three conclusions about these models:

\textbf{\acrshort{SSL} models, in general, improve the reliability.} We observe in both CIFAR-10-C and CIFAR-10.1, almost all \acrshort{SSL} models improve accuracy and uncertainty for in-distribution data except the Context Prediction method. For instance, compared with ERM which has 86.85\%~Accuracy, 0.7136~NLL, and 0.0851~ECE, the Rotation Prediction, Geometric Transformation Prediction (Affine), and Jigsaw Puzzles achieve better performances by around 4\% in terms of Accuracy, lower NLL and ECE than by 0.3 and 0.02 respectively. These improvements are even more significant under the distributional shifts, where with \acrshort{SSL} models performance improves by more than 5\% in Accuracy, 0.9 in NLL, and 0.04 in ECE in CIFAR-10-C and more than 4\% in Accuracy, 0.4 in NLL, and 0.03 in ECE in CIFAR-10.1.

\textbf{\acrshort{SSL} models, however, can hurt the reliability.} In contrast, we observe in background-less images in MNIST-C, \acrshort{SSL} models do not have a significant improvement on in-distribution data. Moreover, although having the same corruption techniques as CIFAR-C, using \acrshort{SSL} even hurt the generalization and uncertainty estimate on out-of-distribution data. In particular, while ERM achieves 1.5439~NLL, 77.64\%~Accuracy, and 0.1759~ECE, all others \acrshort{SSL} degrade these criteria remarkably to less 72\%~Accuracy, more than 1.9~NLL, and 0.2~ECE (except Geometric Transformations Prediction).

\textbf{Jigsaw Puzzle, in general, is the best stable and reliable \acrshort{SSL} method.} Regarding background informative samples in CIFAR-10, \acrshort{SSL} models, notably, the Jigsaw Puzzle performs best in accuracy both on in/out-of-distribution data in CIFAR-10-C and CIFAR-10.1, and even better than Rotation in terms of uncertainty estimation on out-of-distribution data with 2.5242~NLL and 0.2407~ECE in CIFAR-10-C. However, Rotation Prediction is the best uncertainty estimate method on in-distribution data. Regarding MNIST-C, even though all \acrshort{SSL} models hurt reliability, the Jigsaw is still one of the methods that is least impacted negatively under generalization and uncertainty estimation.

\textbf{Distributional Shifts Details.}
\begin{figure}[ht!]
\begin{center}
\includegraphics[width=1.0\linewidth]{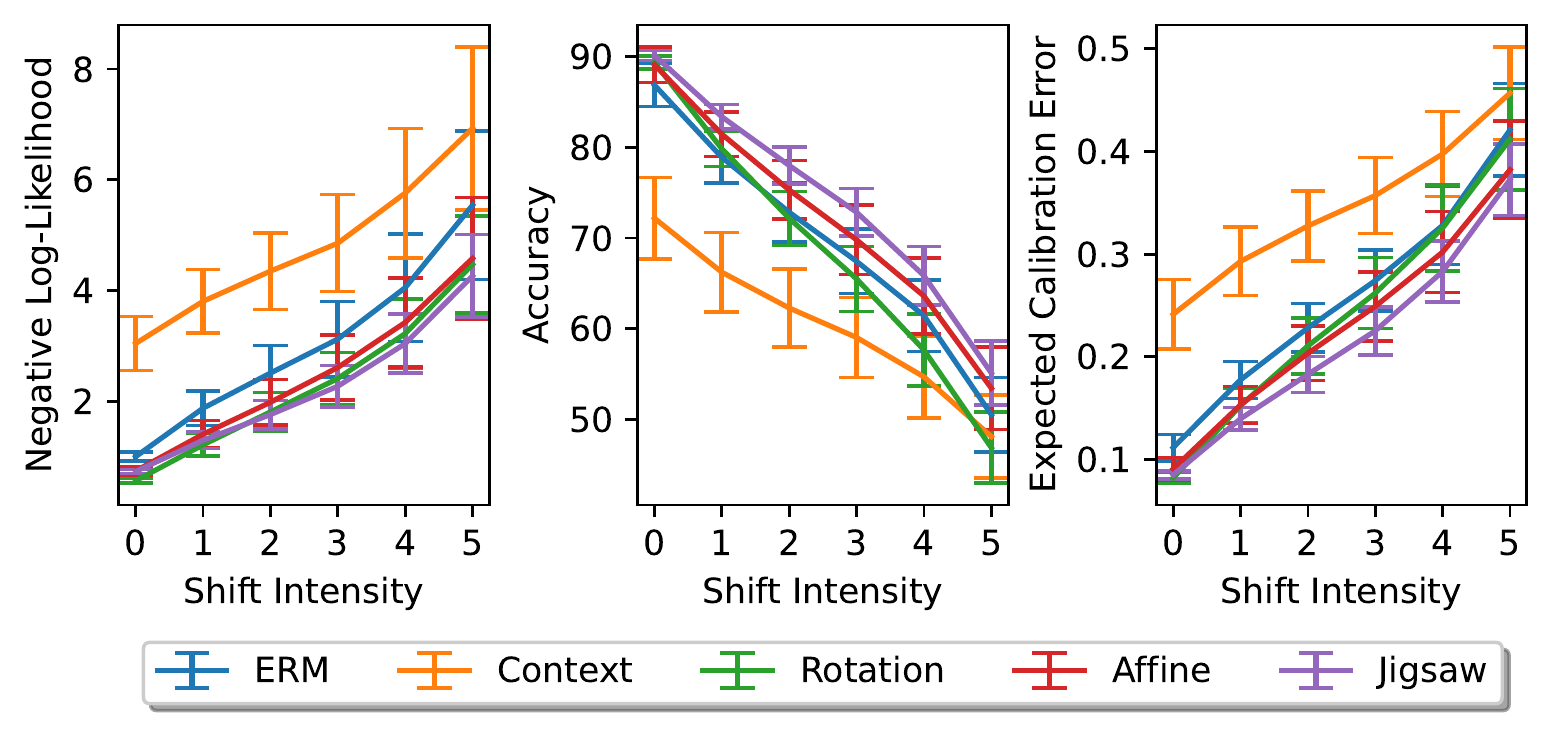}
\end{center}
   \caption{In-Out-of-distribution performance using CIFAR-10-C with Wide Resnet-28-10. We plot NLL, accuracy, and ECE for varying corruption intensities; each result is the mean performance over 10 runs and 19 corruption types. The error bars represent a fraction of the standard deviation across corruption types. Jigsaw (violet) performs best across all metrics (More detail is provided in appendix~Appendix~\ref{apd:results}).}
\label{fig:cifar-skew}
\vskip -0.2in
\end{figure}

To understand more about the advantage of \acrshort{SSL} methods under distributional shifts, we visualize the comparison of NLL, Accuracy, and the ECE on in/out-of-distribution of CIFAR-10-C for varying corruption intensities in the Figure~\ref{fig:cifar-skew} (More detail is provided in the Figure~\ref{fig:dcifar-skew} in Appendix~\ref{apd:results}). From these sub-figures, we observe a consistent performance across different corruption intensities, where Jigsaw is the best method with the lowest NLL, ECE, and highest Accuracy across the in-distribution data, and 5 shift intensity levels. However, we also observe all methods are still impacted by the distributional shifts level, showing that it's not robust under distributional shifts. Specifically, all methods are degraded considerably when the shifting intensity level increases, from around 0.5 to more than 8 in NLL, 90\% to less than 60\% in Accuracy, and 0.05 to around more than 0.3 in ECE between no shift and the level 5 shift intensity respectively.

\textbf{Uncertainty Details.}
\begin{figure}[ht!]
\centering
\subfloat{
  \includegraphics[width=0.2\linewidth]{{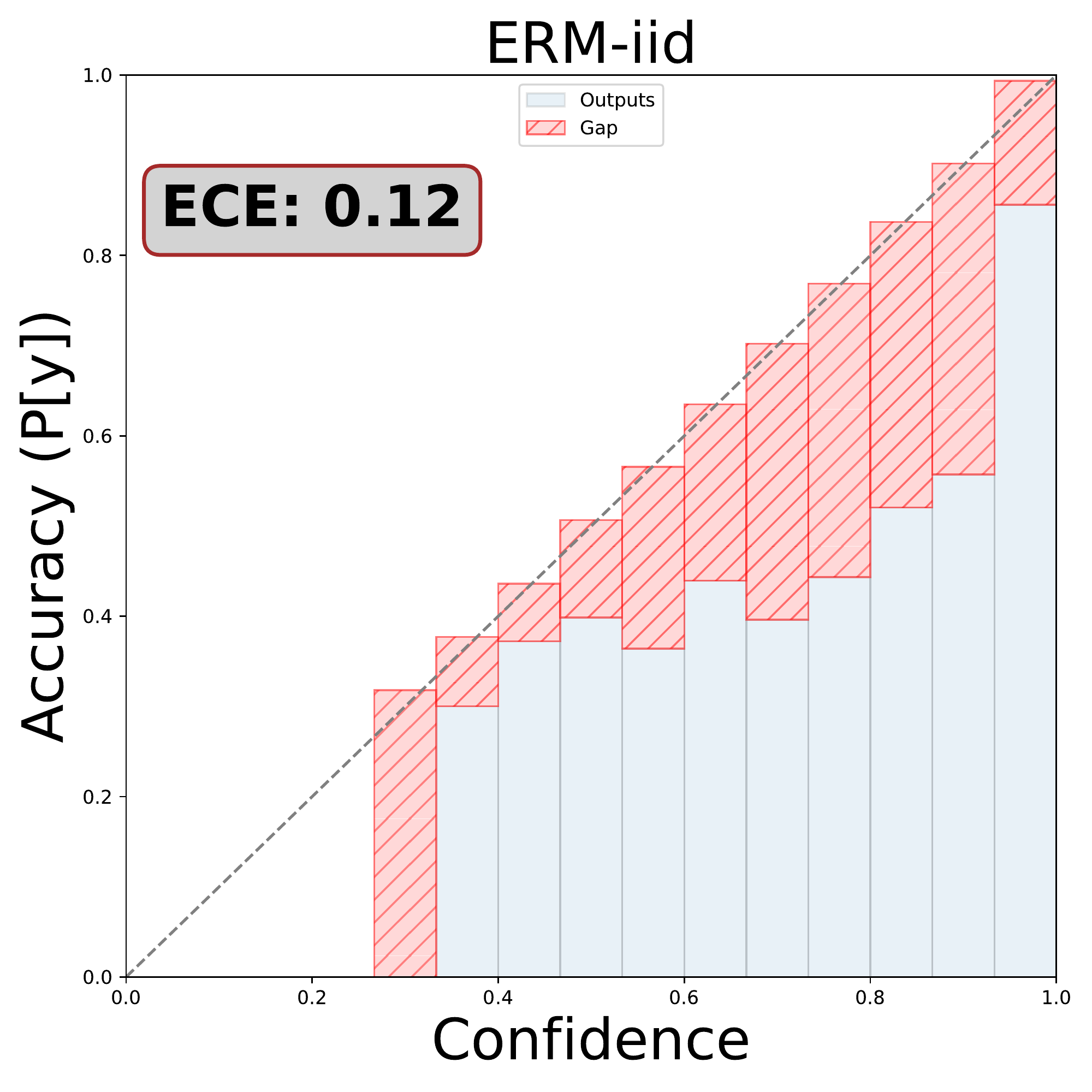}}
}
\hspace{-0.12in}
\subfloat{
  \includegraphics[width=0.2\linewidth]{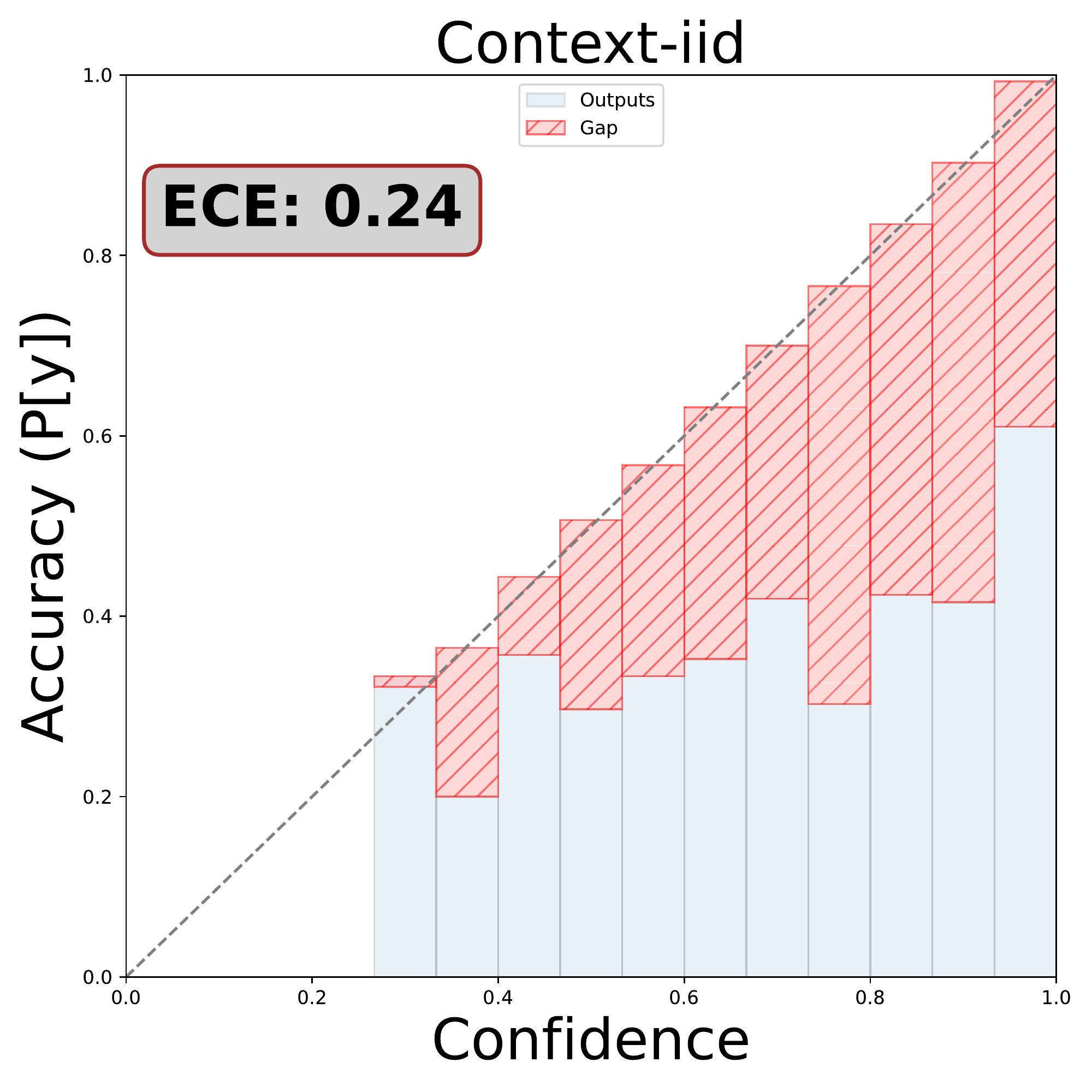}
}
\hspace{-0.12in}
\subfloat{
  \includegraphics[width=0.2\linewidth]{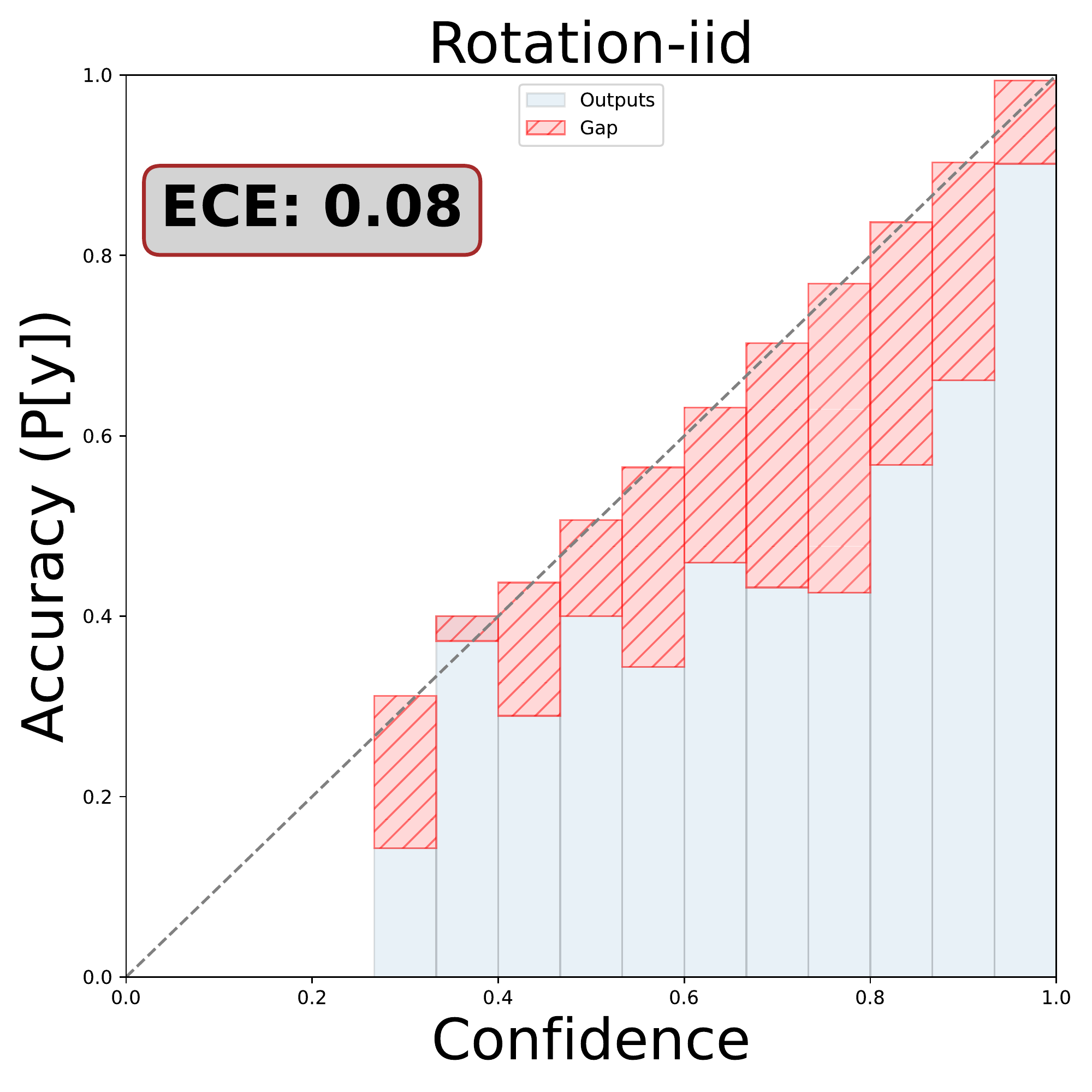}
}
\hspace{-0.12in}
\subfloat{
  \includegraphics[width=0.2\linewidth]{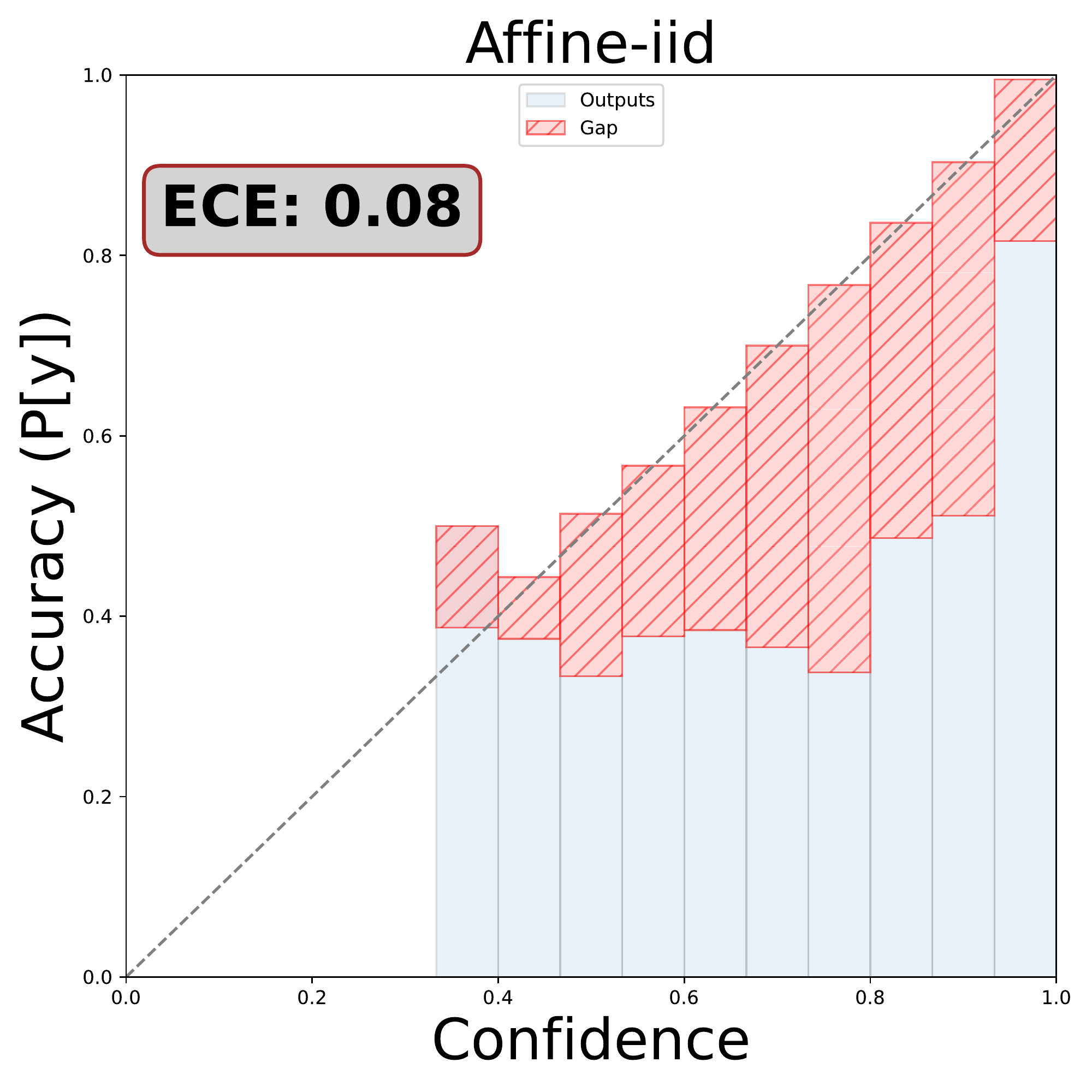}
}
\hspace{-0.12in}
\subfloat{
  \includegraphics[width=0.2\linewidth]{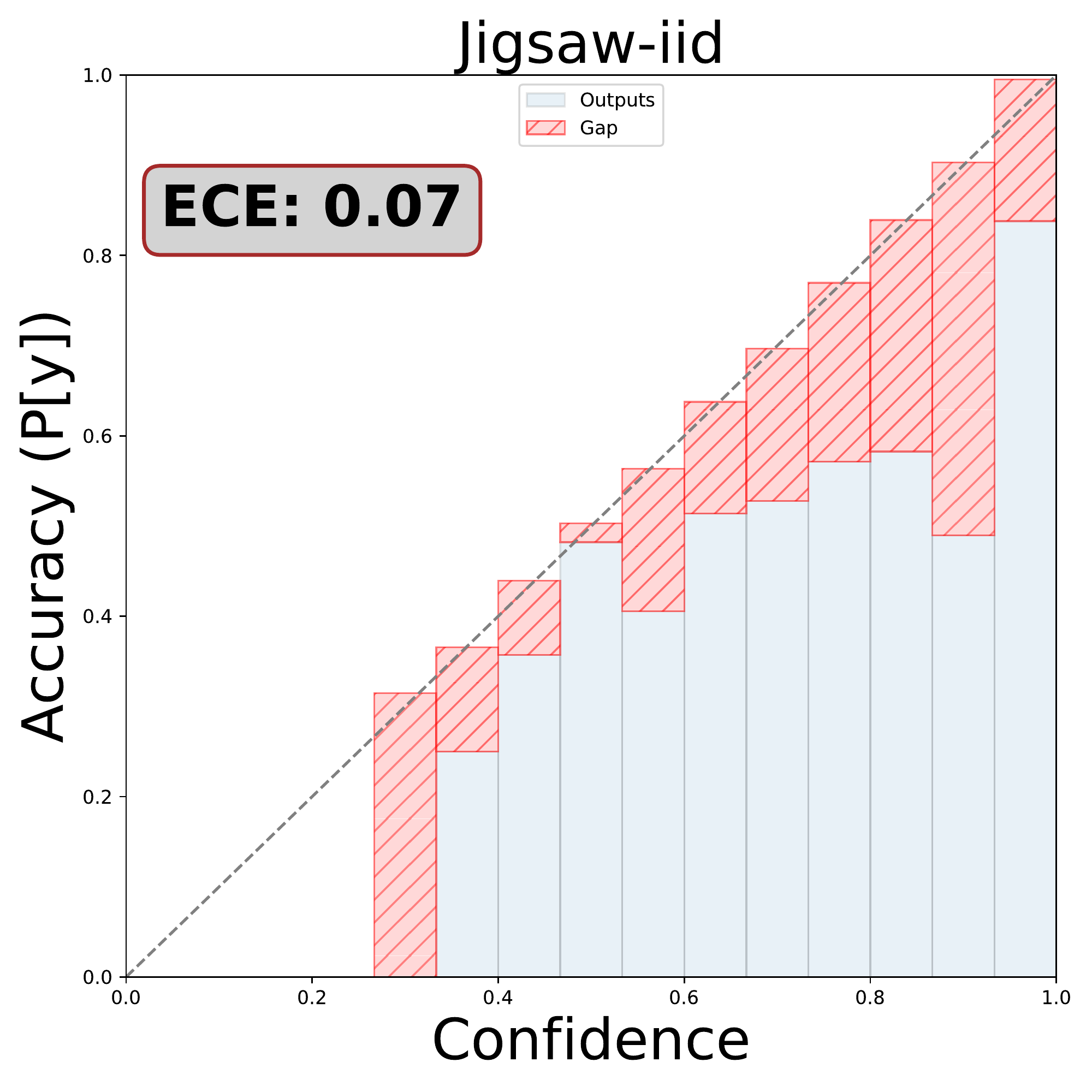}
}
\vspace{-0.12in}
\subfloat{
  \includegraphics[width=0.2\linewidth]{{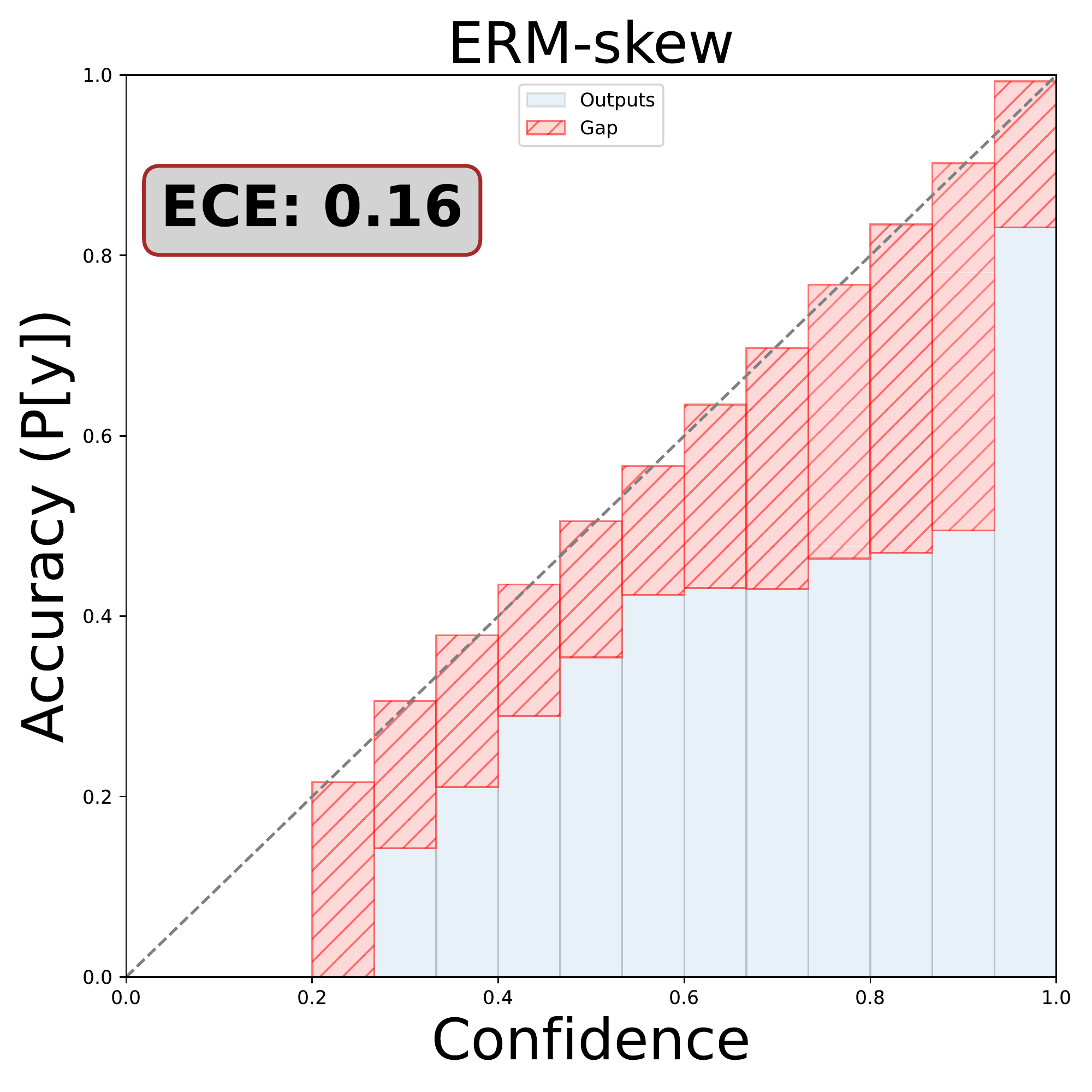}}
}
\hspace{-0.12in}
\subfloat{
  \includegraphics[width=0.2\linewidth]{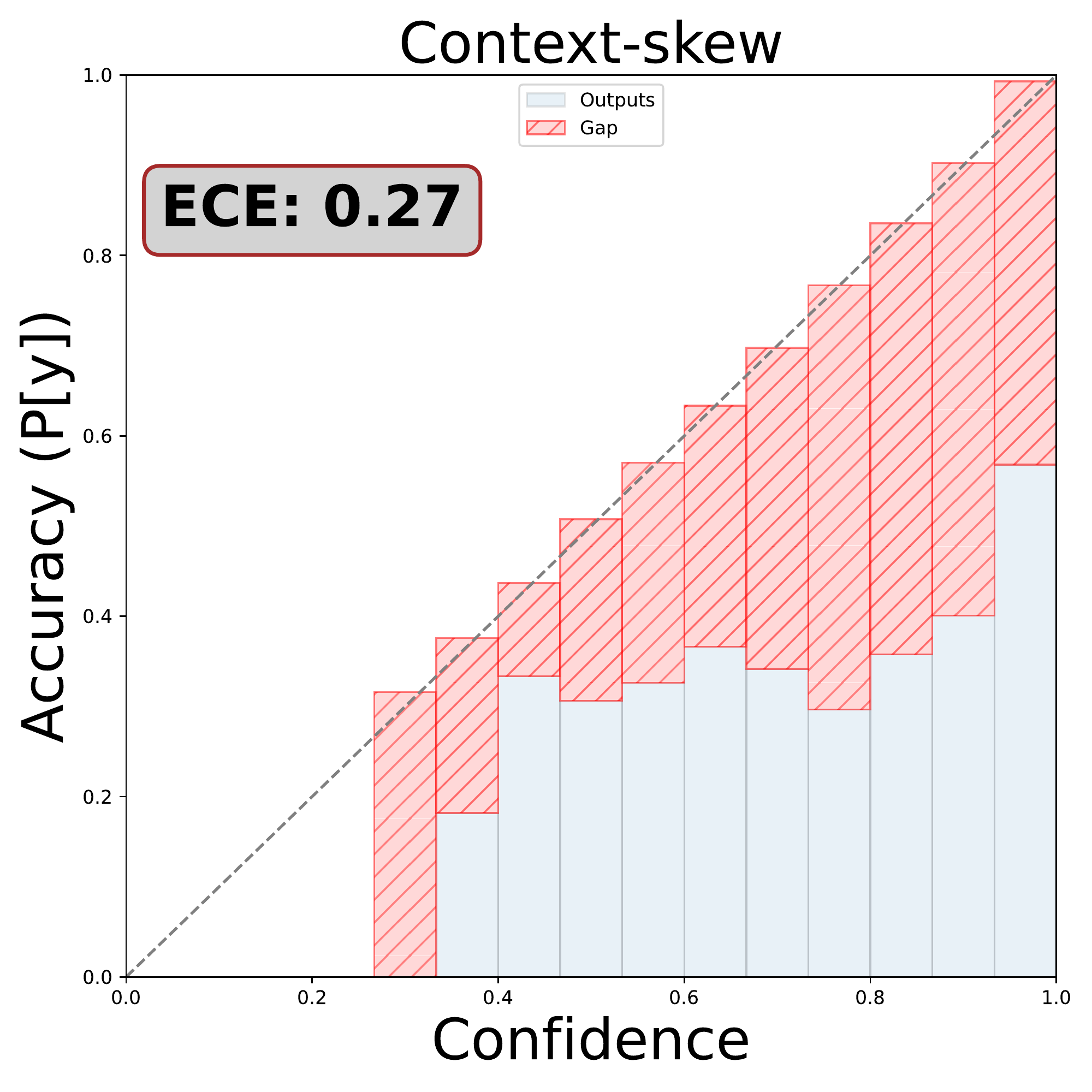}
}
\hspace{-0.12in}
\subfloat{
  \includegraphics[width=0.2\linewidth]{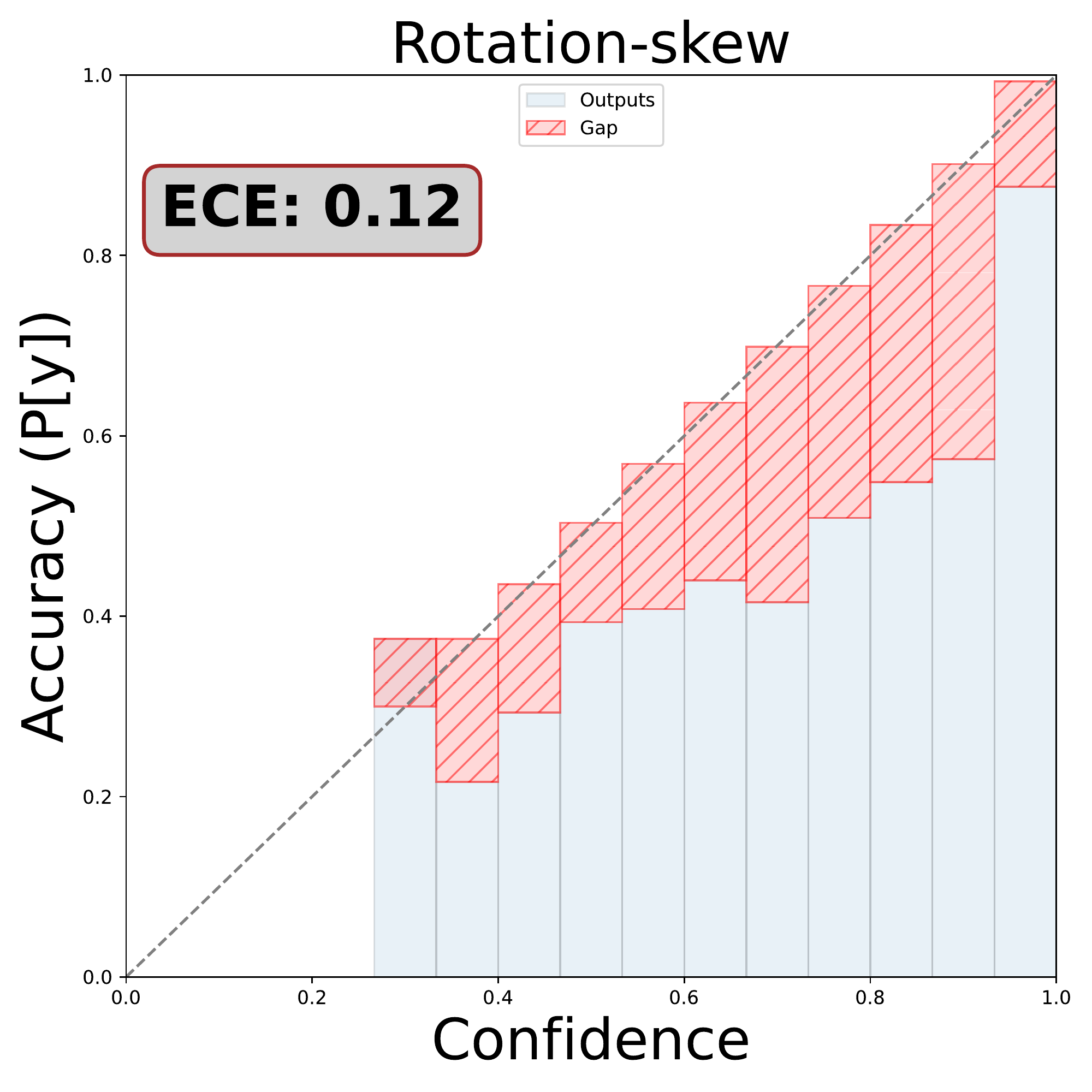}
}
\hspace{-0.12in}
\subfloat{
  \includegraphics[width=0.2\linewidth]{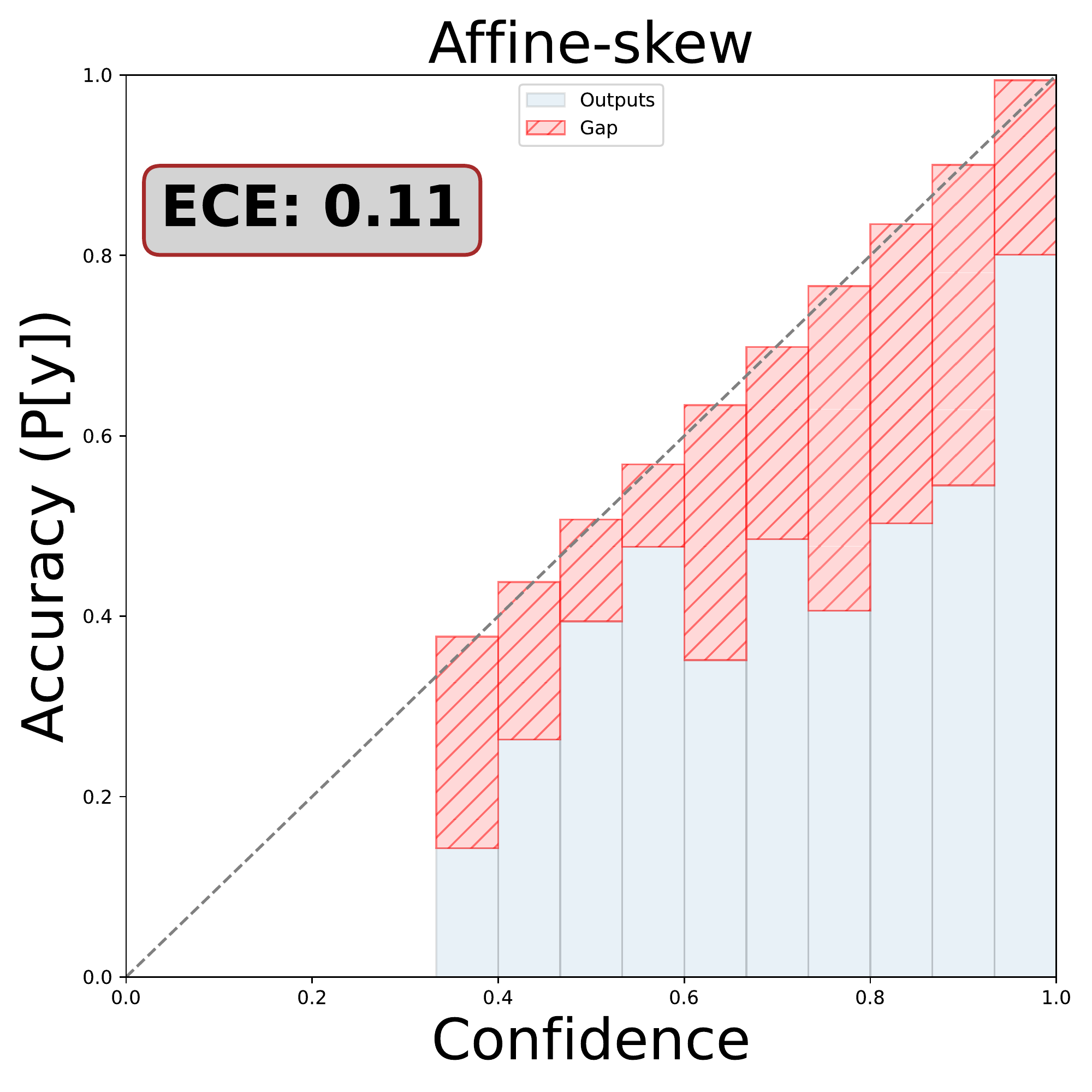}
}
\hspace{-0.12in}
\subfloat{
  \includegraphics[width=0.2\linewidth]{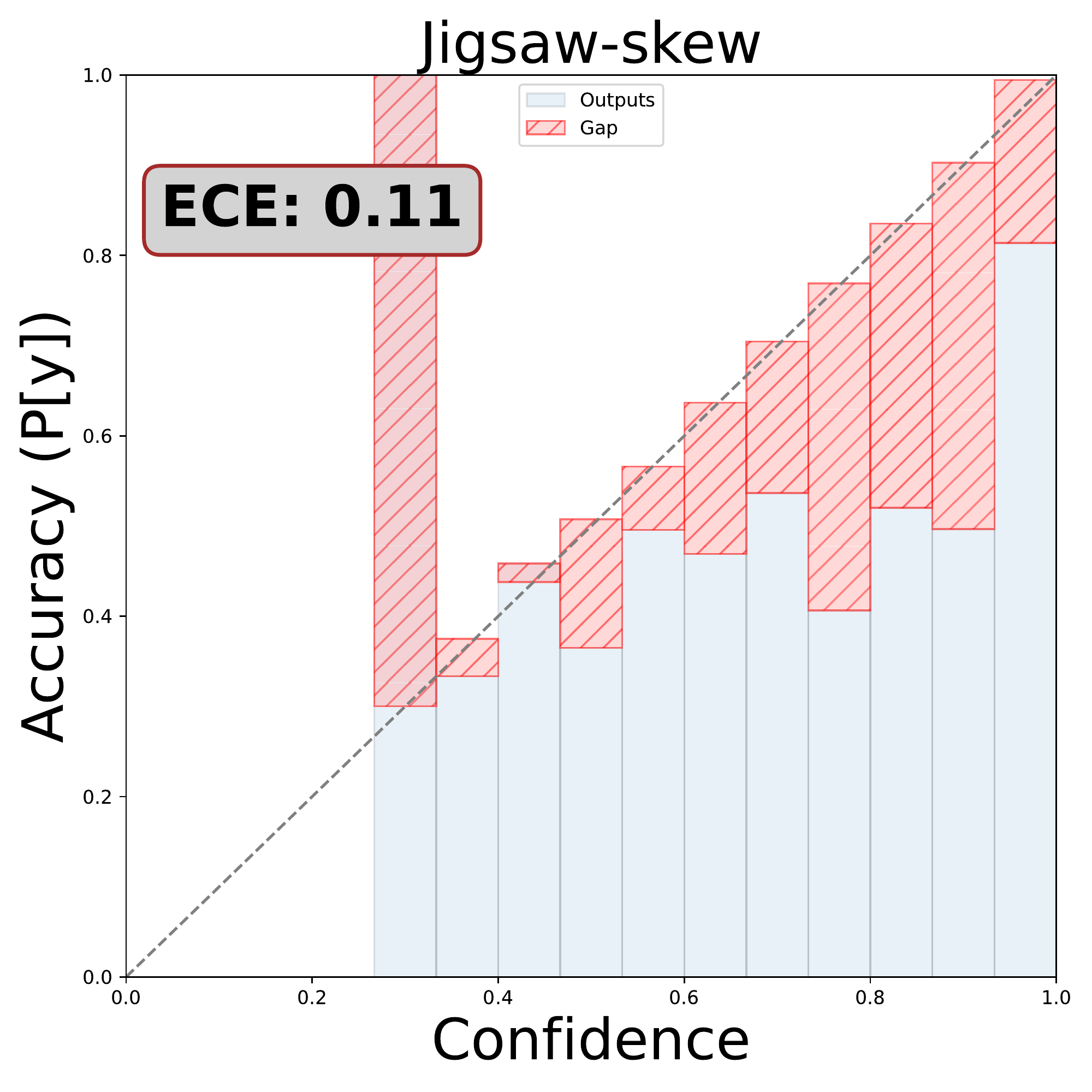}
}
\vspace{-0.12in}
\subfloat{
  \includegraphics[width=0.2\linewidth]{{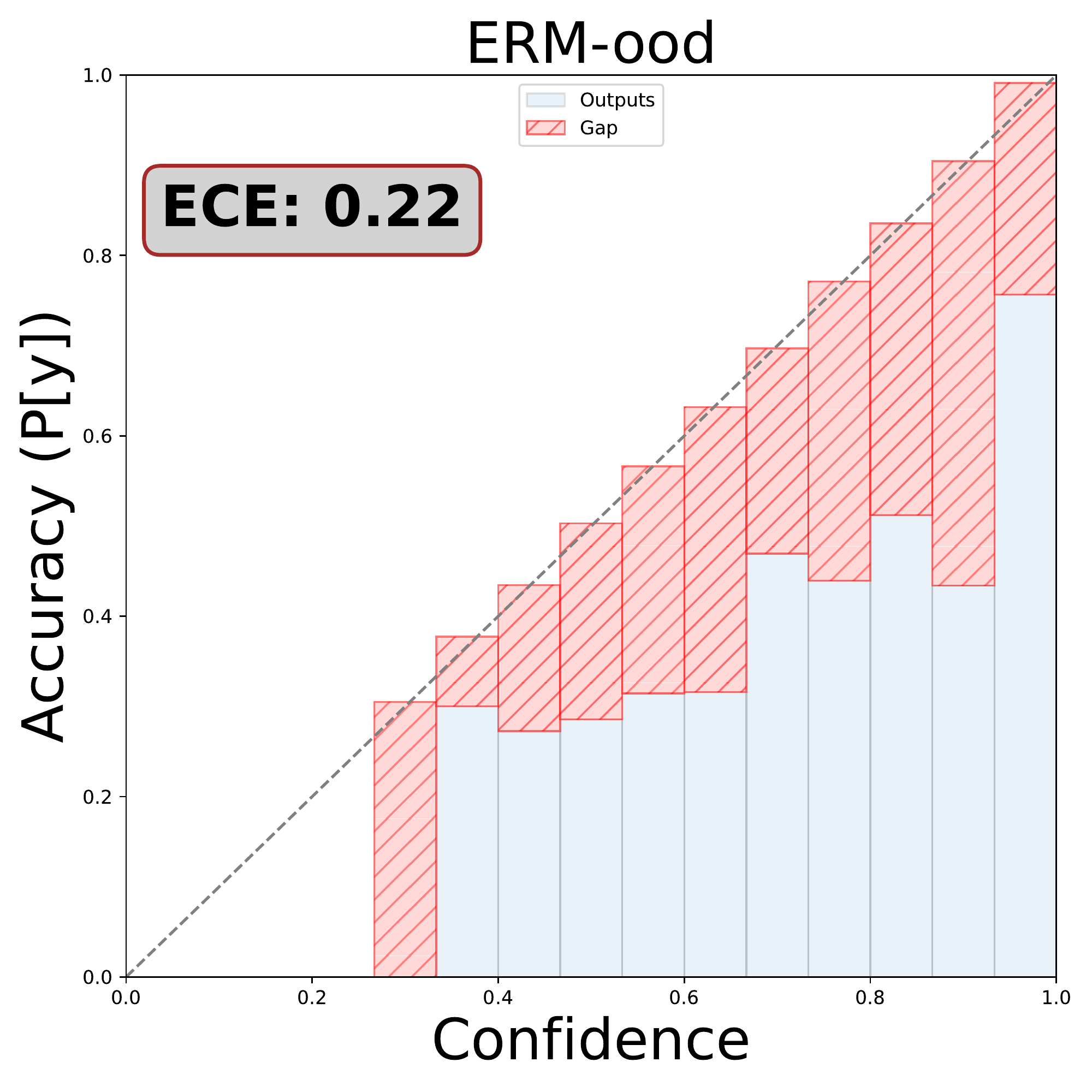}}
}
\hspace{-0.12in}
\subfloat{
  \includegraphics[width=0.2\linewidth]{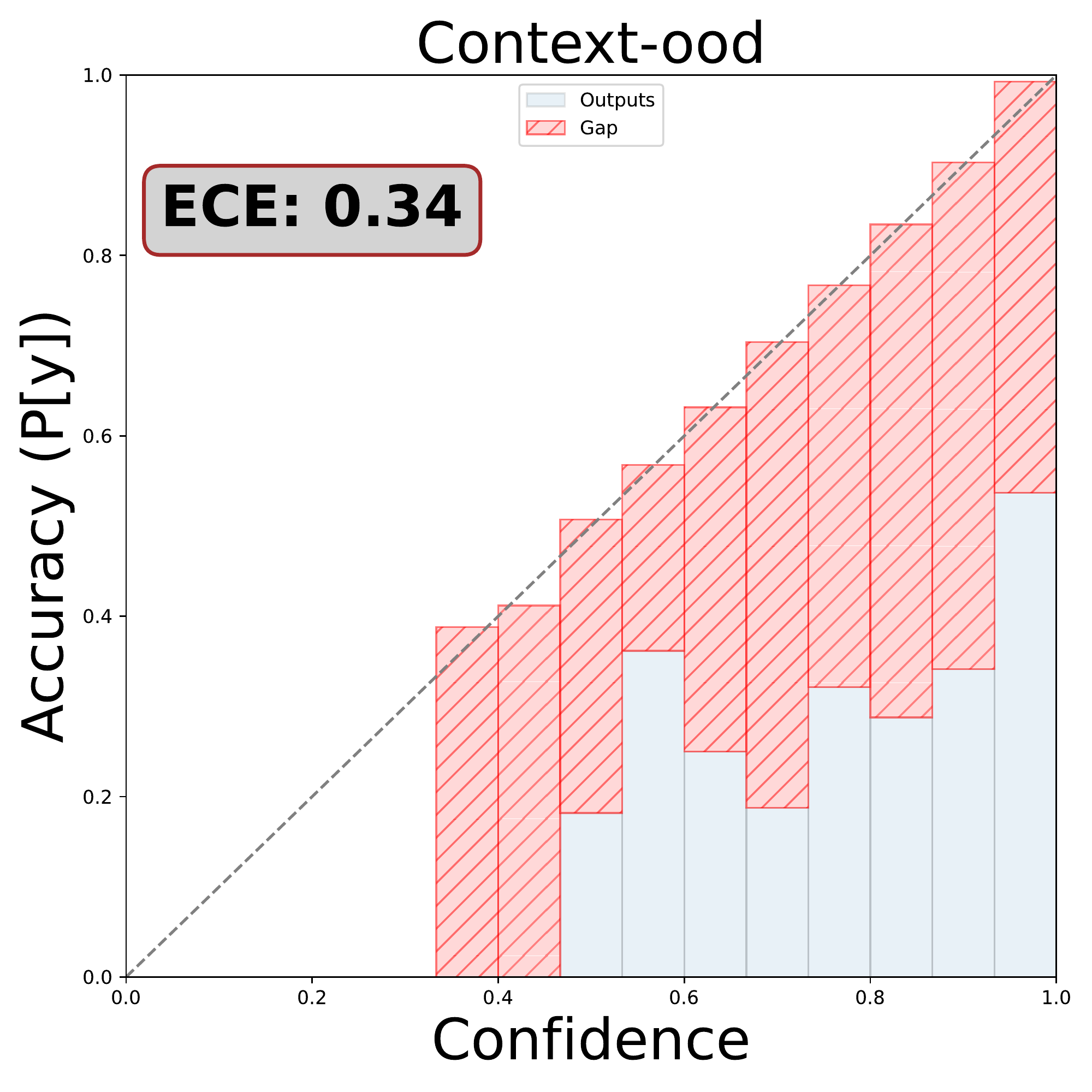}
}
\hspace{-0.12in}
\subfloat{
  \includegraphics[width=0.2\linewidth]{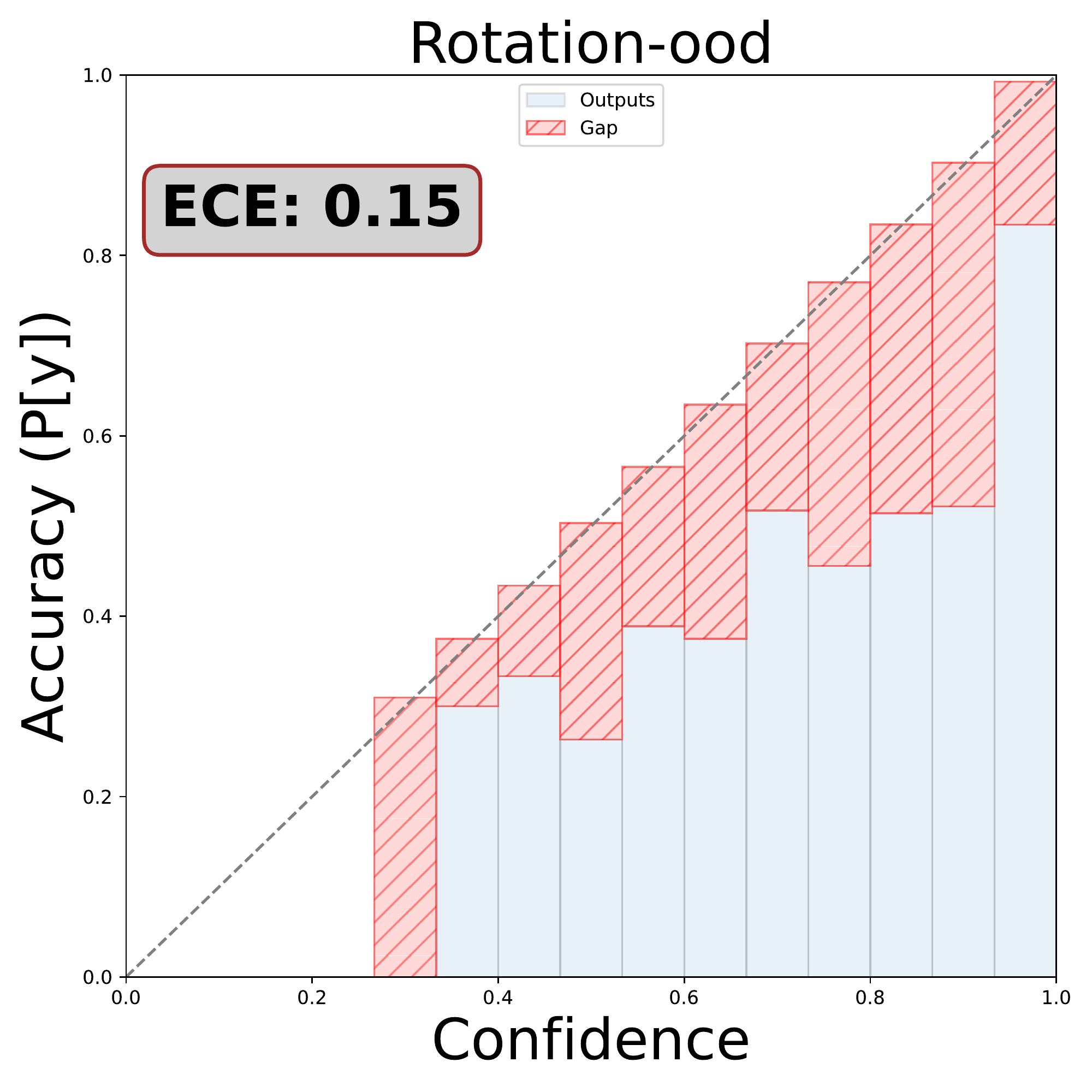}
}
\hspace{-0.12in}
\subfloat{
  \includegraphics[width=0.2\linewidth]{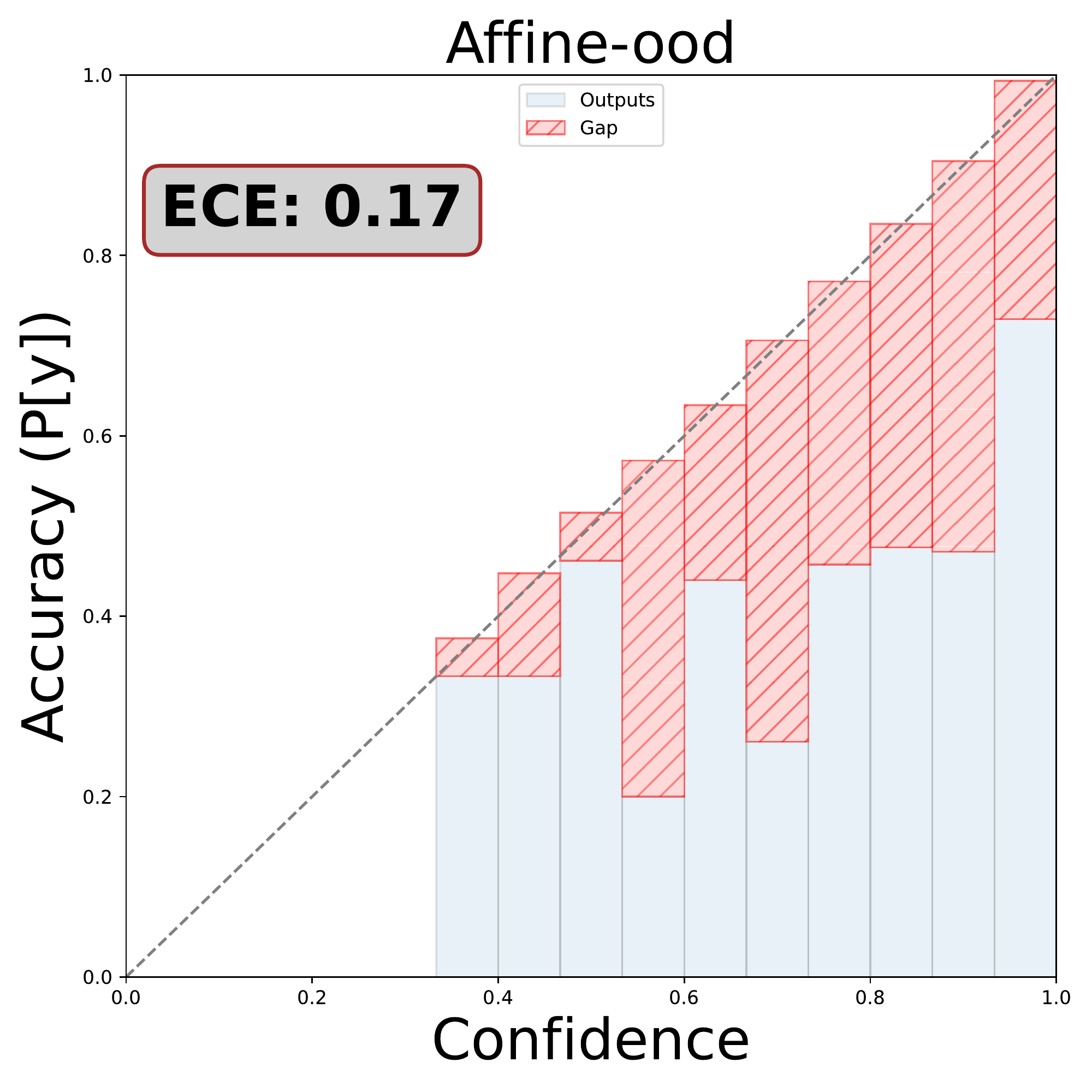}
}
\hspace{-0.12in}
\subfloat{
  \includegraphics[width=0.2\linewidth]{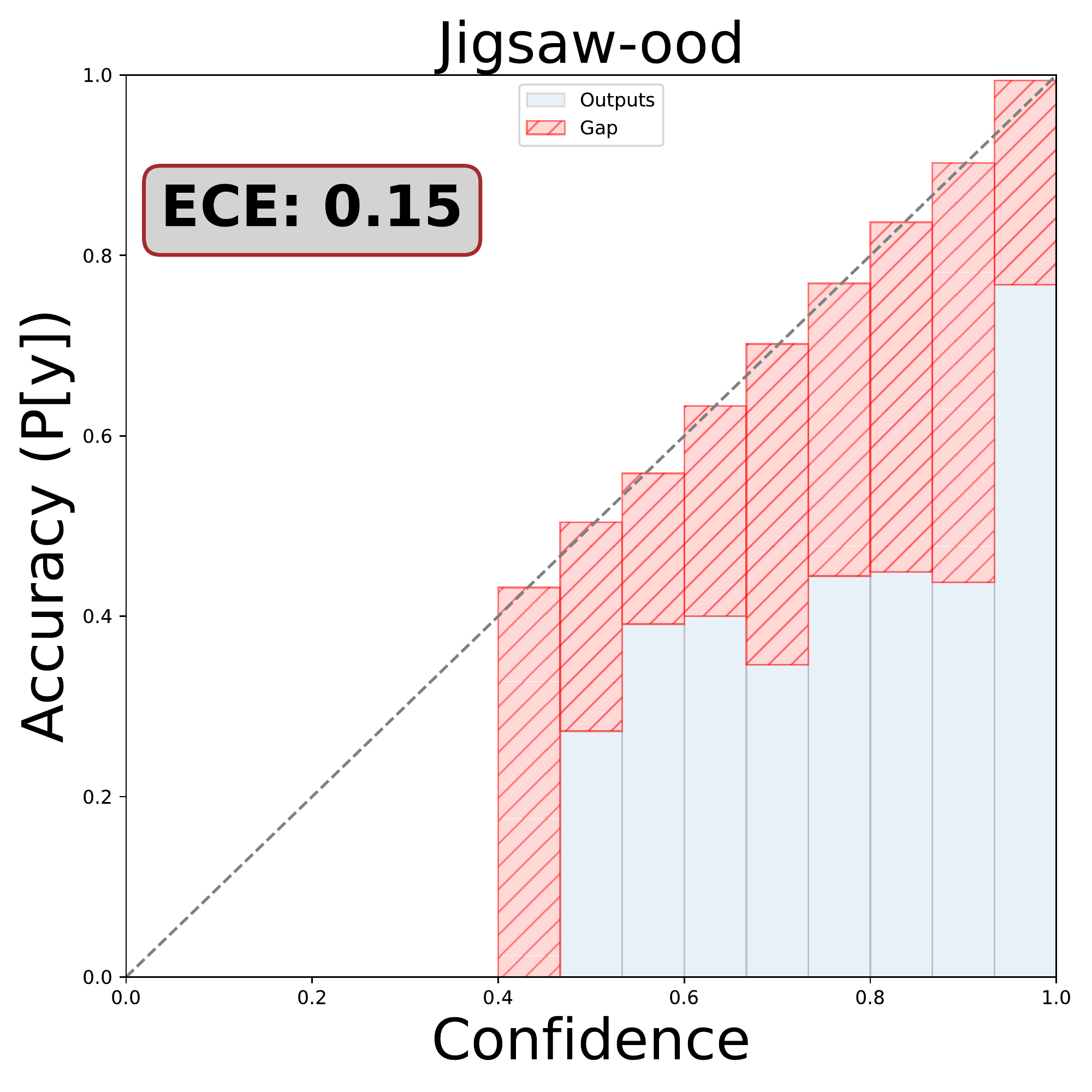}
}
\caption{Reliability diagrams comparison of ECE under in of distribution, the distributional shifts of "brightness" type of corruptions on CIFAR-10, and the distributional shifts of CIFAR-10.1 v-6.}
\label{fig:reliability-diagram}
\vskip -0.2in
\end{figure}

\begin{figure}[ht!]
\vspace{-0.2in}
\hspace{-0.2in}
\subfloat{
  \includegraphics[width=0.34\linewidth]{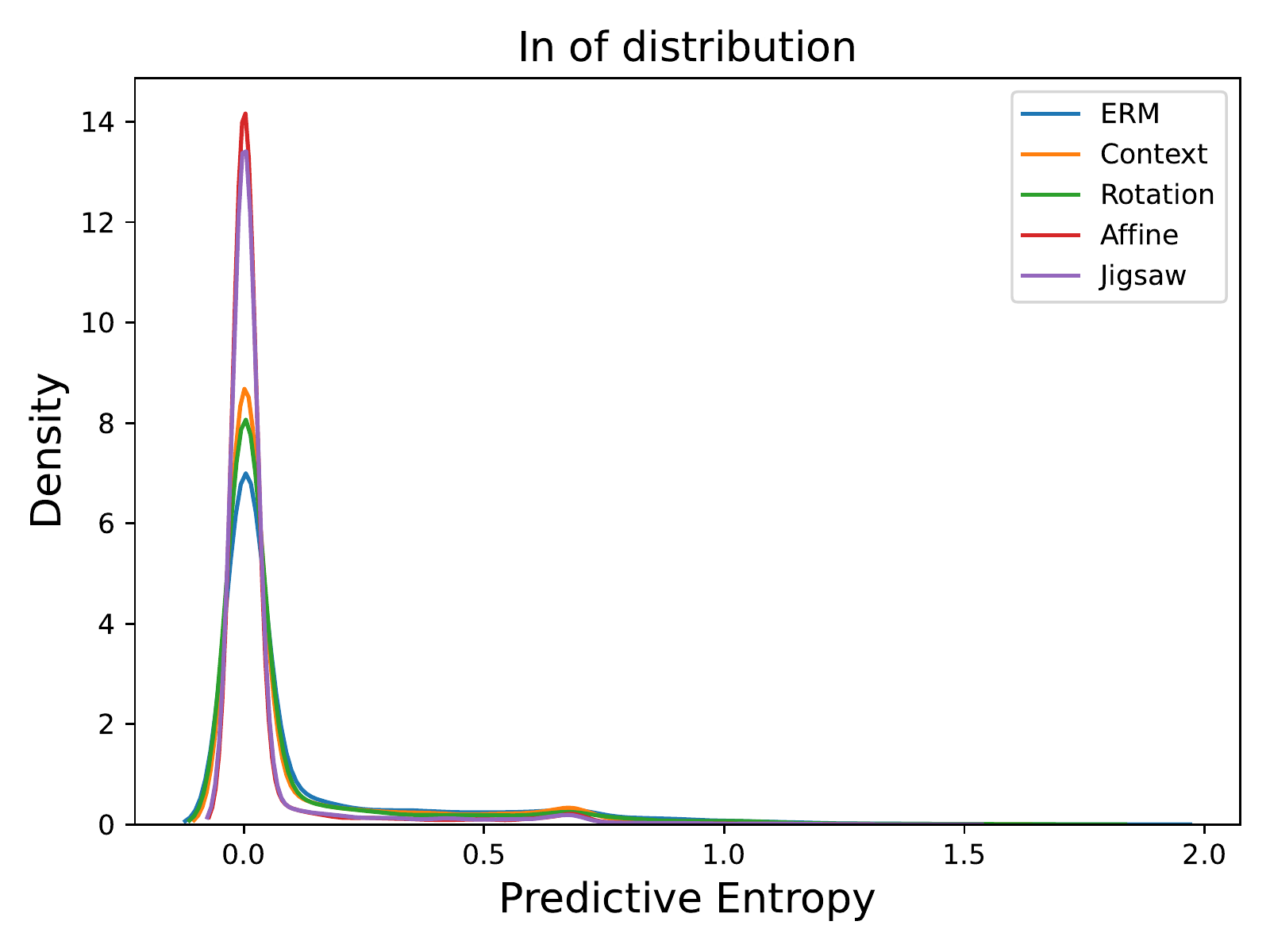}
}
\hspace{-0.15in}
\subfloat{
  \includegraphics[width=0.34\linewidth]{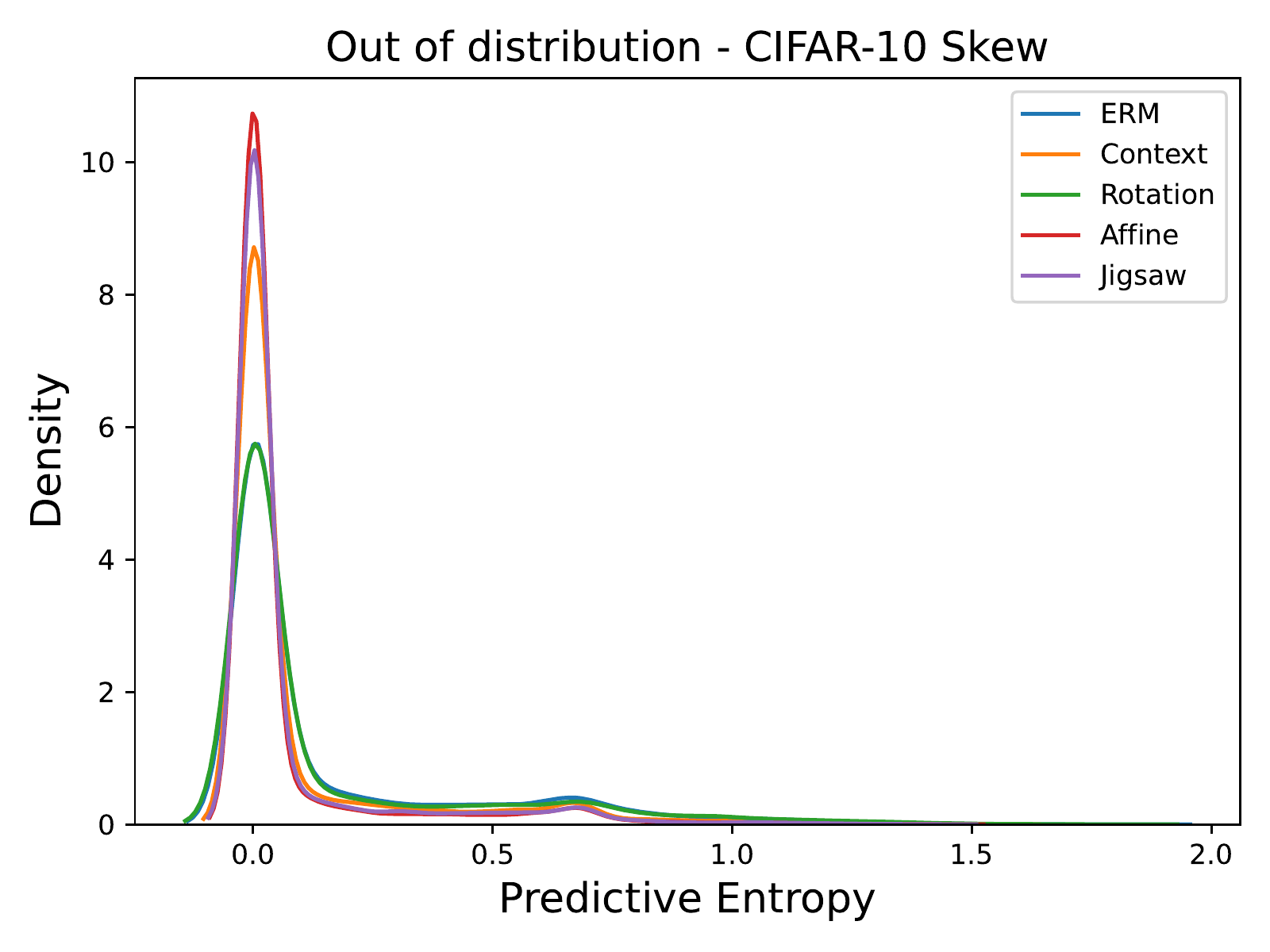}
}
\hspace{-0.15in}
\subfloat{
  \includegraphics[width=0.34\linewidth]{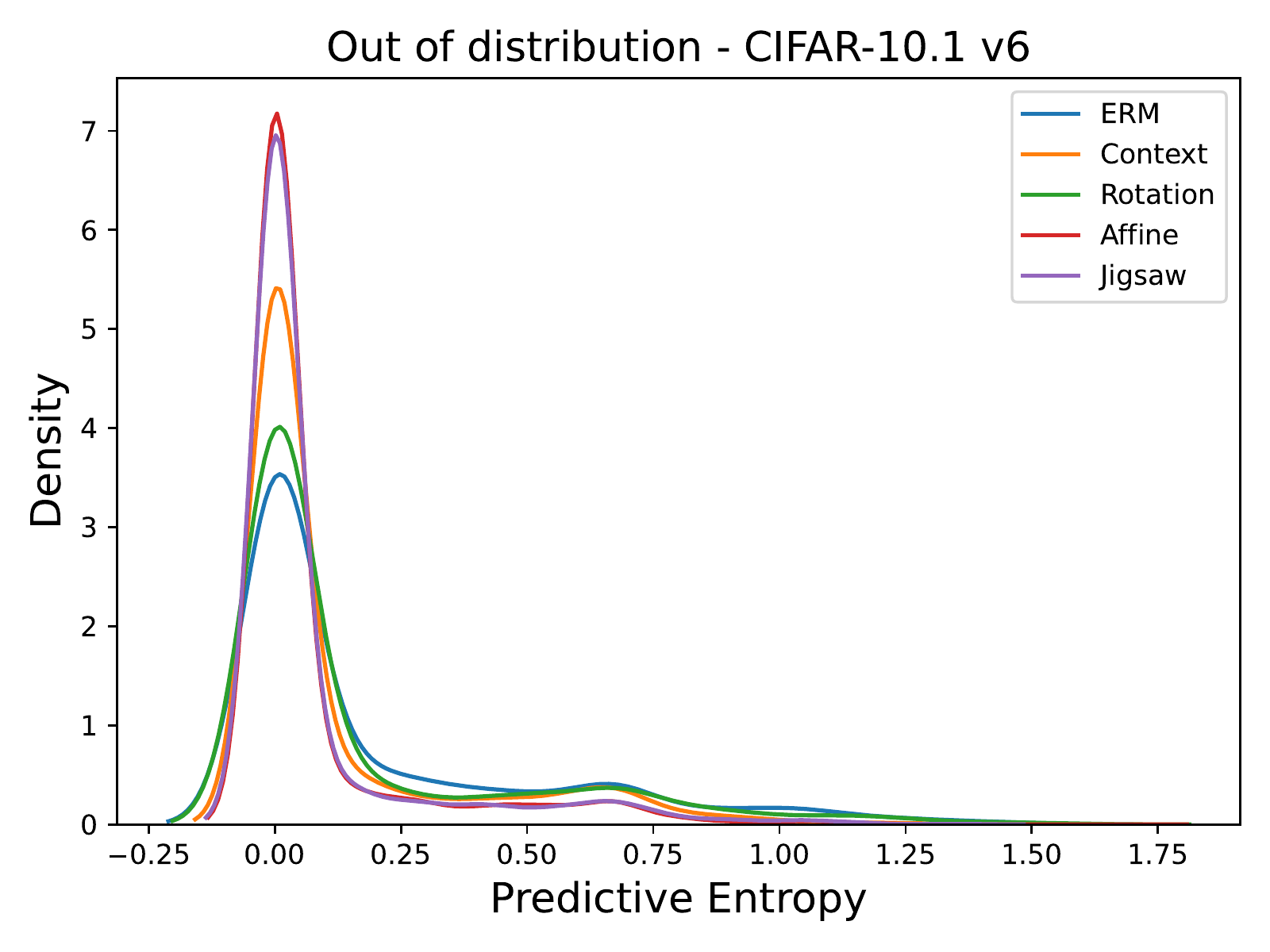}
}
\caption{Histograms of predictive entropy for Wider Resnet-28-10 architecture trained on CIFAR10 dataset and test on in of distribution, out of distribution with CIFAR-10 Skew by "brightness" type, and the real-world out of distribution with CIFAR-10.1~v6.}
\label{fig:PDF-diagram}
\end{figure}

The lower NLL and ECE of \acrshort{SSL} than ERM in Table~\ref{tab:results} shows a better uncertainty estimate. However, due to the increase in accuracy at the same time, this uncertainty improvement may simply be because of a side effect of good representation. To confirm whether \acrshort{SSL} can actually improve predictive uncertainty, we visualize the reliability diagram of each method with in-of-distribution, distributional shift with the "brightness" skew (an instance of corruption in CIFAR-10-C), and the real-world distributional shifts with class-balanced in CIFAR-10.1 v6 in Figure~\ref{fig:reliability-diagram}. Comparing ERM and Jigsaw Puzzle which is the most stable method, we observe although the ECE of Jigsaw is lower in three tests, Jigsaw also looks less over-confidence than ERM. However, it still can not be calibrated well, and sometimes suffer from under-confidence in the skew test. The under-confidence can be observed more clearly in Figure~\ref{fig:PDF-diagram} (and Figure~\ref{fig:dPDF-diagram} in Appendix~\ref{apd:results}). The predictive entropy of Jigsaw is often 0, which mean does not provide any informative information about the distributional shifts, showing it is unable to recognize the out-of-distribution data. From these observations, we can conclude that although \acrshort{SSL} can improve the uncertainty estimate, it does not provide a good uncertainty estimate enough for a good reliability model.

\subsubsection{Language Task}
\begin{table}[ht!]
\caption{Results for BERT and GPT2 on MNLI showing negative log-likelihood (lower is better), accuracy (higher is better), and expected calibration error (lower is better).}
\label{tab:NLPresults}
\centering
\scalebox{0.95}{
\begin{tabular}{lcccccc}
\toprule
\textbf{Model} & \textbf{NLL($\downarrow$)} & \textbf{Accuracy($\uparrow$)} & \textbf{ECE($\downarrow$)} & \textbf{OOD NLL($\downarrow$)} & \textbf{OOD Acc($\uparrow$)} & \textbf{OOD ECE($\downarrow$)}\\
\midrule
BERT~\cite{devlin-etal-2019-bert} & \textbf{.5309} & \textbf{.8242} & .1546 & \textbf{.5138} & \textbf{.8103} & .1661\\
GPT2~\cite{radford2019language} & .9205 & .5332 & \textbf{.04625} & 1.0897 & .4872 & \textbf{.031}\\
\bottomrule
\end{tabular}}
\end{table}

Table \ref{tab:NLPresults} summarizes the results on the natural language inference dataset MultiNLI comparing the language models BERT and GPT2 on Negative Log Likelihood (NLL), Accuracy, and Expected Calibration Error (ECE). These results show that BERT generalizes fairly well to the mild dataset shift, but we still see an average of a 1.4\% decrease in accuracy and .012 increase in ECE when presented with out-of-distribution (OOD) data. Interestingly, the generalization properties of BERT are better on OOD data, with the loss lowering by .023. We also see that GPT2 does not perform or generalize very well to the dataset shift with accuracy drops of 4.6\% and increase in loss by .1692. However, GPT2 is much better calibrated than BERT with ECE scores of .04625 and .031 on the in/out-of-distribution data respectively. 

%% file: paragraphs/5_conclusion.tex
\section{Conclusion}
Despite many people being aware of the importance of \acrshort{SSL} in generalization, there has been little investigation into the uncertainty estimation of \acrshort{SSL} methods. To the best of our knowledge, our work provides the first fair comparison of both generalization and uncertainty under distributional shifts for \acrshort{SSL} models in both vision and language tasks. For the vision tasks, the \acrshort{SSL} variants, including: Context, Rotation, Geometric Transformation Prediction, and Jigsaw Puzzles are deployed as an auxiliary task, then compared under standard MNIST and CIFAR-10 distributional shifts datasets. Our experimental results demonstrate \acrshort{SSL} can improve reliability, but not always, and often only good in generalization. The results also suggest that Jigsaw Puzzle performs very well and may be the best stable method across \acrshort{SSL} family. For the language tasks, we fine-tune BERT and GPT2 on the MultiNLI dataset where the spoken transcriptions represent a distributional shift. 
We find that BERT has a better performance overall and does a superior job of generalizing with the out-of-distribution data than does GPT2. However, GPT2 outperforms BERT regarding calibration. With the main motivation to provide a good benchmark for reliable \acrshort{SSL} models, we hope our initial results, including the source code, experiments, and observations are helpful and a good starting point for new \acrshort{SSL} methods in Reliable Machine Learning.

%% file: appendix/app_main.tex
\newpage
\appendix
\section*{\centering\textbf{Benchmark for Uncertainty \& Robustness in Self-supervised Learning (Supplementary Material)}}

In this supplementary material, we collect the remaining materials that were deferred from the main paper. In particular, we provide additional information about our experiments, including: sufficient details about the dataset in Appendix~\ref{apd:dataset}, implementation details to reproduce our experiments in Appendix~\ref{apd:implementation}, further analysis of the results in Appendix~\ref{apd:results}.

\section{Dataset details}\label{apd:dataset}

This appendix provides more detail about the dataset in the main paper, including 3 image datasets widely used for classification tasks in distributional shifts:
\begin{itemize}
\item{\textbf{MNIST-C~\cite{mu2019mnist}}} includes $10,000$ binary images of dimension $(1, 28, 28)$ in classification problem with 10 classes, generated from the test images of MNIST~\cite{mnist} over 15 corruptions noise type $d\in$ \big\{identity, shot noise, impulse noise, glass blur, motion blur, shear, scale, rotate, brightness, translate, stripe, fog, spatter, dotted line, zigzag, candy edges\big\}.
\item{\textbf{CIFAR-10-C~\cite{hendrycks2018benchmarking}}} contains $10,000$ colored samples of dimension $(3, 32, 32)$ with 10 classes in classification problem. The distributional shifts data are generated from the test images of CIFAR-10~\cite{cifar10} over a total of 19 corruptions, with 15 standard noise type $d\in$ \big\{identity, gaussian noise, shot noise, impulse noise, defocus blue, frosted glass blur, motion blur, zoom blur, snow, frost, fog, brightness, contrast, elastic, pixelate, jpeg\big\} and 4 more extra noise type \big\{gaussian blur, saturate, spatter, speckle noise\big\}. For each noise corruption, there are 5 skew intensities represented for the level noise to evaluate the robustness, so we have a total of 95 out-of-distribution test sets.
\item{\textbf{CIFAR-10.1~\cite{recht2018cifar10.1}}} has $2,021$ images for version v4 and $2,000$ images with class balanced for version v6, of dimension $(3, 32, 32)$ in classification problem with the same 10 classes in CIFAR-10. This dataset is collated in real-world images on the Internet and is a subset of the Tiny Images~\cite{torralba2008tinyimages}. This dataset is used to additionally test the reliability of models under real-world distributional shifts.
\end{itemize}

\section{Implementation details}\label{apd:implementation}

\begin{table}[ht!]
\caption{Condition architectures, hyper-parameters, and their default values in our experiments.}
\label{tab:hyper-params}
\centering
\scalebox{0.9}{
\begin{tabular}{lll}
\toprule
\textbf{Datset} & \textbf{Hyper-parameters} & \textbf{Default value} \\
\midrule
\multirow{10}{*}{MNIST} & backbone & LeNet5\\
  & features dims & 84\\
  & optimizer & SGD(momentum = 0.9, nesterov = True)\\
  & learning rate & 0.1\\ 
  & batch size & 128\\ 
  & iterations & 10000\\
  & decay iterations & [3000, 6000, 9000], gamma = 0.2\\
  & step eval & 500\\
  & \acrshort{SSL} weight & 0.1\\
  & ERM~\cite{abraham1983statistical} & input dims = $(1, 32, 32)$\\
  & Context~\cite{contextprediction} & class = \{0,.., 8\}, input dims = $(1, 33, 33)$\\
  & Rotation~\cite{TransformNet} & class = [0, 45, 90, 135, 180, 225, 270, 315], input dims = $(1, 32, 32)$\\
  & Affine~\cite{TransformNet} & rotation = [0, 180], scaling = [0.7, 1.3], \\ & & shearing = [0.3, -0.3], transition = [0, 2, -2], input dims = $(1, 32, 32)$\\
  & \acrshort{SSL} Jigsaw~\cite{carlucci2019domain} & permutations = 30, class =  \{0,.., 30\}, input dims = $(1, 33, 33)$\\
\midrule
\multirow{10}{*}{CIFAR-10} & backbone & Wide Resnet 28-10\\
  & features dims & 640\\
  & optimizer & SGD(momentum = 0.9, nesterov = True)\\
  & learning rate & 0.1\\ 
  & batch size & 128\\ 
  & iterations & 78125\\
  & decay iterations & [23437, 46875, 62500], gamma = 0.2\\
  & step eval & 500\\
  & \acrshort{SSL} weight & 0.1\\
  & ERM~\cite{abraham1983statistical} & input dims = $(3, 32, 32)$\\
  & Context~\cite{contextprediction} & class = \{0,.., 8\}, input dims = $(3, 33, 33)$\\
  & Rotation~\cite{TransformNet} & class = [0, 45, 90, 135, 180, 225, 270, 315], input dims = $(3, 32, 32)$\\
  & Affine~\cite{TransformNet} & rotation = [0, 180], scaling = [0.7, 1.3], \\ & & shearing = [0.3, -0.3], transition = [0, 2, -2], input dims = $(3, 32, 32)$\\
  & \acrshort{SSL} Jigsaw~\cite{carlucci2019domain} & permutations = 30, class =  \{0,.., 30\}, input dims = $(3, 33, 33)$\\
\bottomrule
\end{tabular}}
\vspace{-0.2in}
\end{table}

In this appendix, we describe the data-processing techniques, neural network architectures, hyper-parameters, and details for reproducing our experiments.

\textbf{Data processing techniques.}
For experiments related to the MNIST dataset, we receive an image with input size 28 × 28 x 1 pixels, then do scaling to 32 × 32 x 1 or 33 × 33 x 1 pixels to fit with LeNet5 backbone, depending on each \acrshort{SSL} methods in Table~\ref{tab:hyper-params}. The ideal size is 32 x 32, however, for Context Prediction and Jigsaw Puzzle, their \acrshort{SSL} technique requires grids that split from input images equally, as a result, for 3x3 grids, we need to scale the original image to 33 x 33. Similarly in CIFAR-10, the only difference is the original input size of CIFAR-10 is 32 x 32 x 3 pixels, so we only need to scale to 32 x 33 for the Context Prediction and Jigsaw Puzzle. We also use the standard normalization for the input sample in CIFAR-10 with the mean is $(0.4914, 0.4822, 0.4465)$ and the standard deviation is $(0.2470, 0.2435, 0.2616)$. \textit{It is worth noting that we do not use any data-augmentation techniques because it may affect the fair comparison between \acrshort{SSL} techniques. For example, using random rotation augmentation in ERM will not be fair with Rotation Prediction because both will rotate the original image.}

\textbf{Architectures and hyper-parameters.}
We list the details of the backbone network and the value of hyper-parameters used for each dataset in Table~\ref{tab:hyper-params}. For experiments related to the MNIST dataset, we use the LeNet5 backbone. For the CIFAR-10 datasets, we use Wide Resnet-28-10. We optimize all models using SGD optimizer and employ the cross-validation set technique for model selection. In particular, for all datasets, we run the test ten times with ten different seeds. For each random seed, we randomly split training and validation from the original training set into 80\% and 20\% splits. and choose the model maximizing the accuracy on the validation set, then compute performance on the test sets after fixed iterations.

\textbf{Dataset, source code, and computing system.}
The source code is provided in the mentioned GitHub in the main paper, including scripts to download the dataset, setup for environment configuration, and our provided code (detail in README.md). We run the code on a single GPU: NVIDIA RTX A5000-24564MiB with 8 CPUs: AMD Ryzen Threadripper 3960X 24-Core, RAM: 32GB, and require 20GB available disk space for storage.

\section{Empirical result details}\label{apd:results}
In this appendix, we show our full results and explain them in more detail when compared with different baseline \acrshort{SSL} methods in each distributional shifts level in the CIFAR-10-C benchmark dataset. Then, we do an analysis to confirm whether \acrshort{SSL} actually helps in terms of uncertainty estimation.

\begin{figure}[ht!]
\subfloat{
  \includegraphics[width=1.0\linewidth]{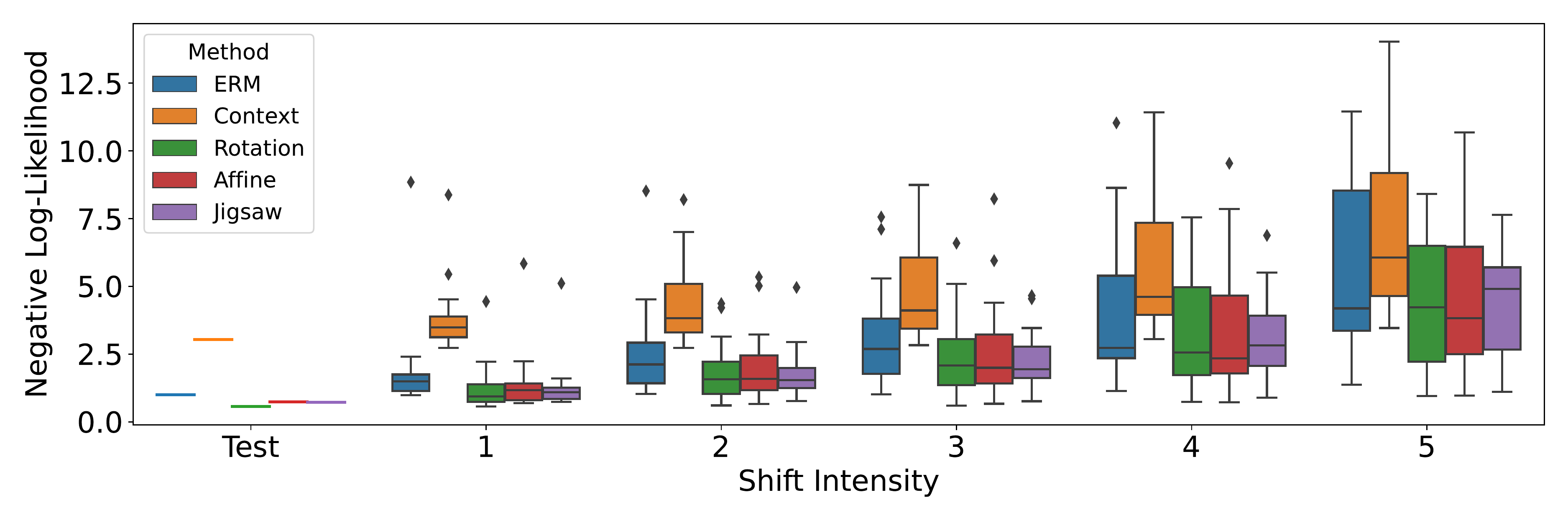}
}
\vspace{-0.2in}
\subfloat{
  \includegraphics[width=1.0\linewidth]{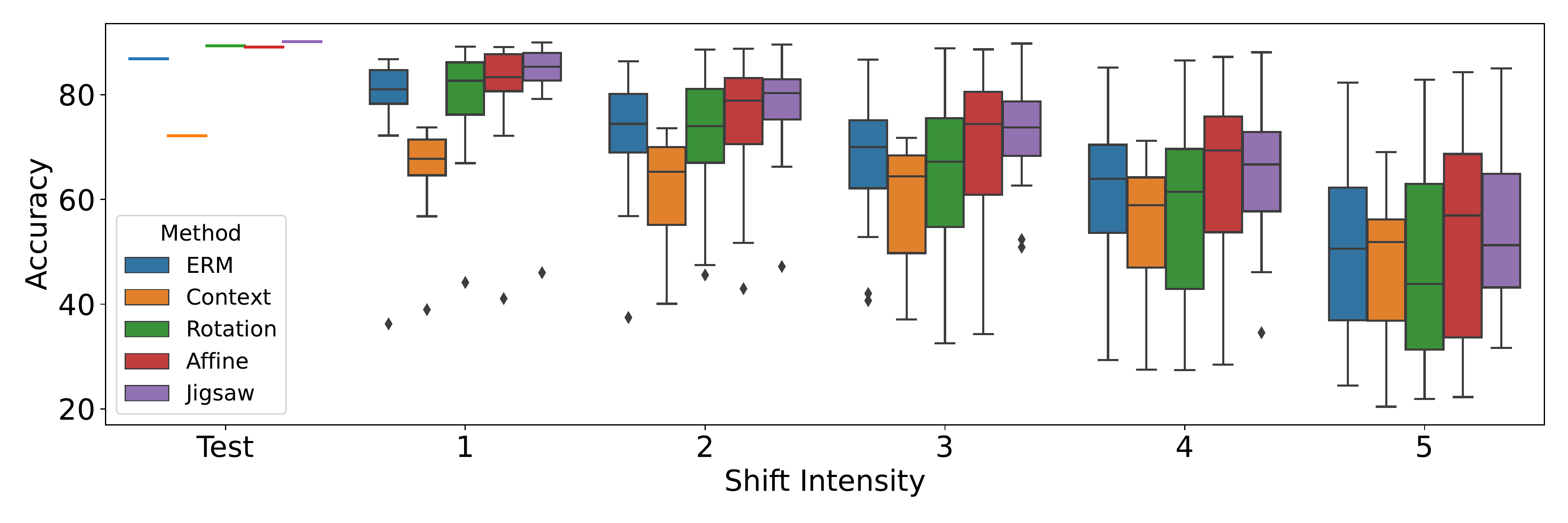}
}
\vspace{-0.2in}
\subfloat{
  \includegraphics[width=1.0\linewidth]{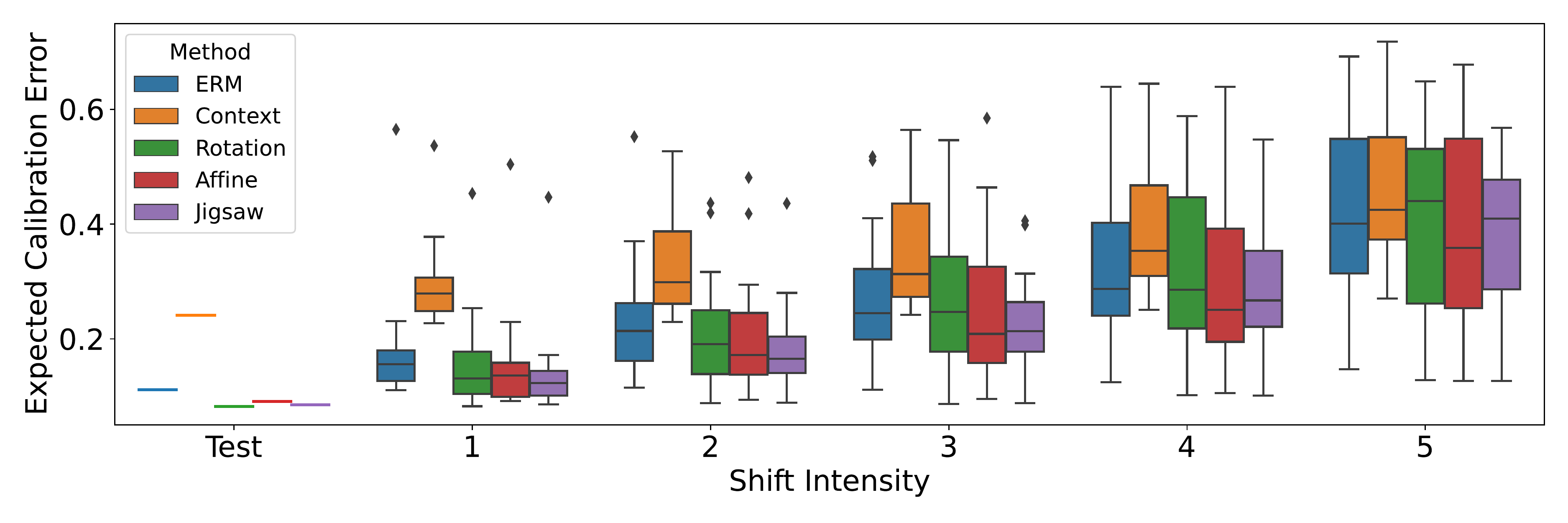}
}
\caption{Comparison under the distributional shift of negative log-likelihood, accuracy, and ECE under all types of corruptions on CIFAR-10. For each method, we show the mean on the test set and summarize the results on each intensity of shift with a box plot. Each box shows the quartiles summarizing the results across all (19) types of shift while the error bars indicate the min and max across different shift types.}
\label{fig:dcifar-skew}
\vspace{-0.2in}
\end{figure}

\textbf{Reliability under distributional shifts by shift-intensity of CIFAR-10-C.} Figure~\ref{fig:dcifar-skew} compares performance of \acrshort{SSL} methods under 5 different shift-intensity levels in CIFAR-10-C. It is worth noting that the statistical comparison between this figure with Figure~\ref{fig:cifar-skew} is different. Figure~\ref{fig:cifar-skew} illustrates the overall statistic of 10 runs and 19 corruption types. Meanwhile, Figure~\ref{fig:dcifar-skew} compares individual statistics of 19 corruption types only. In particular, each box plot shows the minimum, the maximum, the sample median, and the first and third quartiles of 19 different corruption statistics for each \acrshort{SSL} method.\\
The box plot provides consistent performance across methods between shift-intensity levels. The Jigsaw Puzzle is still the best method with the lowest NLL, ECE, and highest Accuracy with a small range of statistics. The results also show degradation in all methods for all criteria when the shift-intensity level increase. Although degradation implies there are no methods that are robust with respect to shift intensity, Jigsaw Puzzle is still observed as the most robust with less degradation amount in all criteria, confirm for the hypothesis that Jigsaw Puzzle is the best stable and reliable \acrshort{SSL} method in Section~\ref{sec:experiments}.

\begin{figure}[ht!]
\hspace{-0.1in}
\subfloat{
  \includegraphics[width=0.208\linewidth]{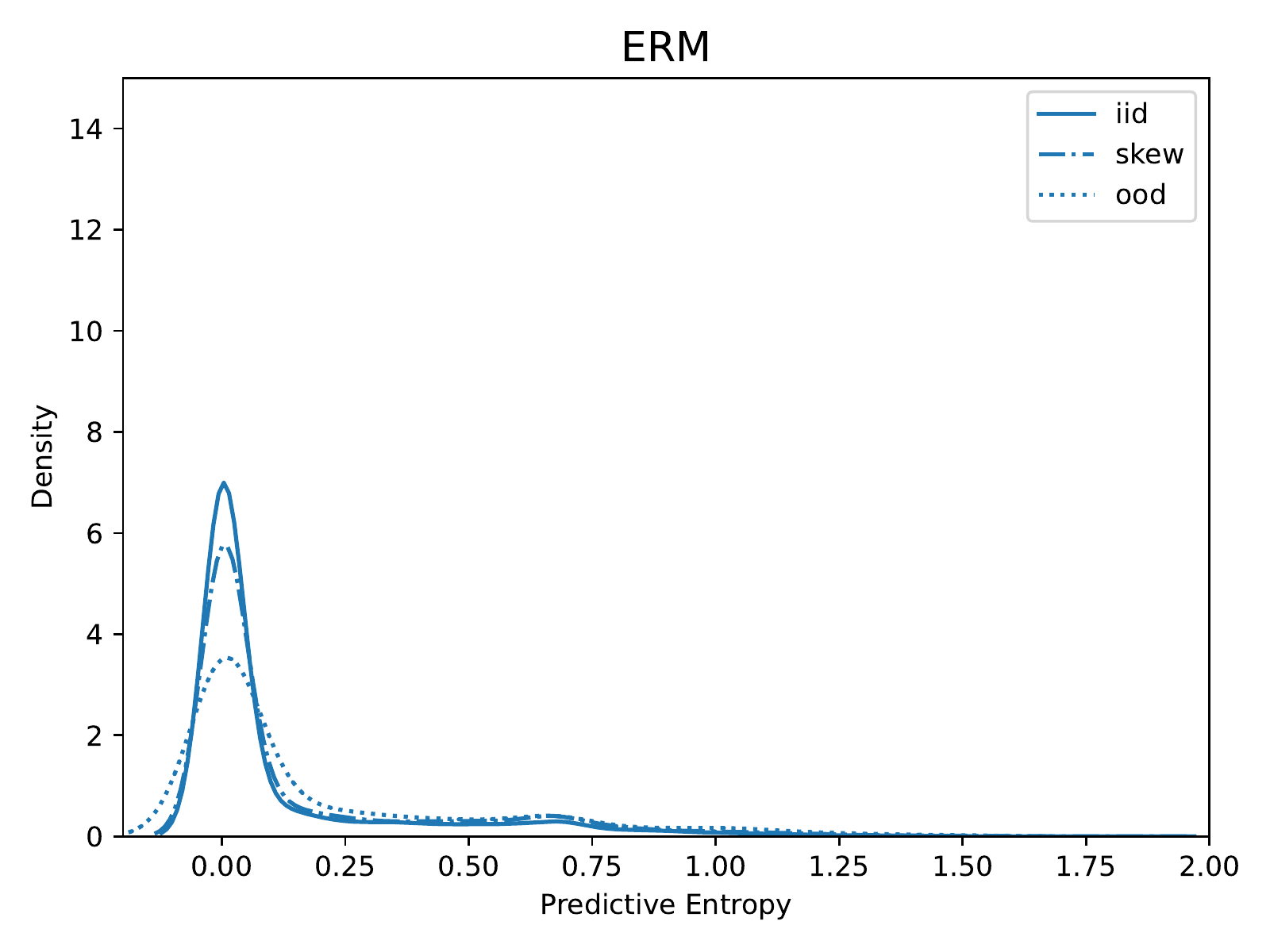}
}
\hspace{-0.15in}
\subfloat{
  \includegraphics[width=0.208\linewidth]{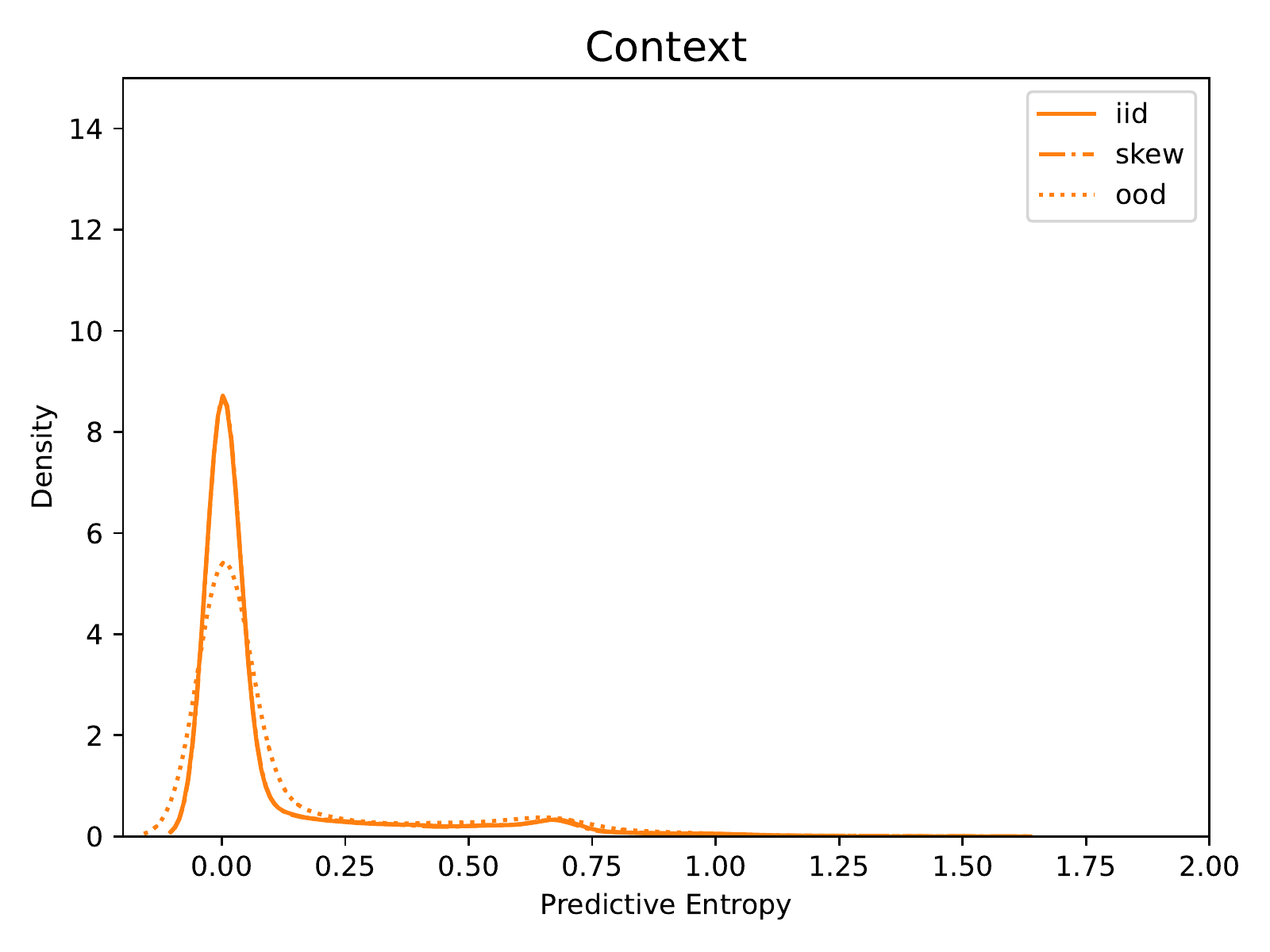}
}
\hspace{-0.15in}
\subfloat{
  \includegraphics[width=0.208\linewidth]{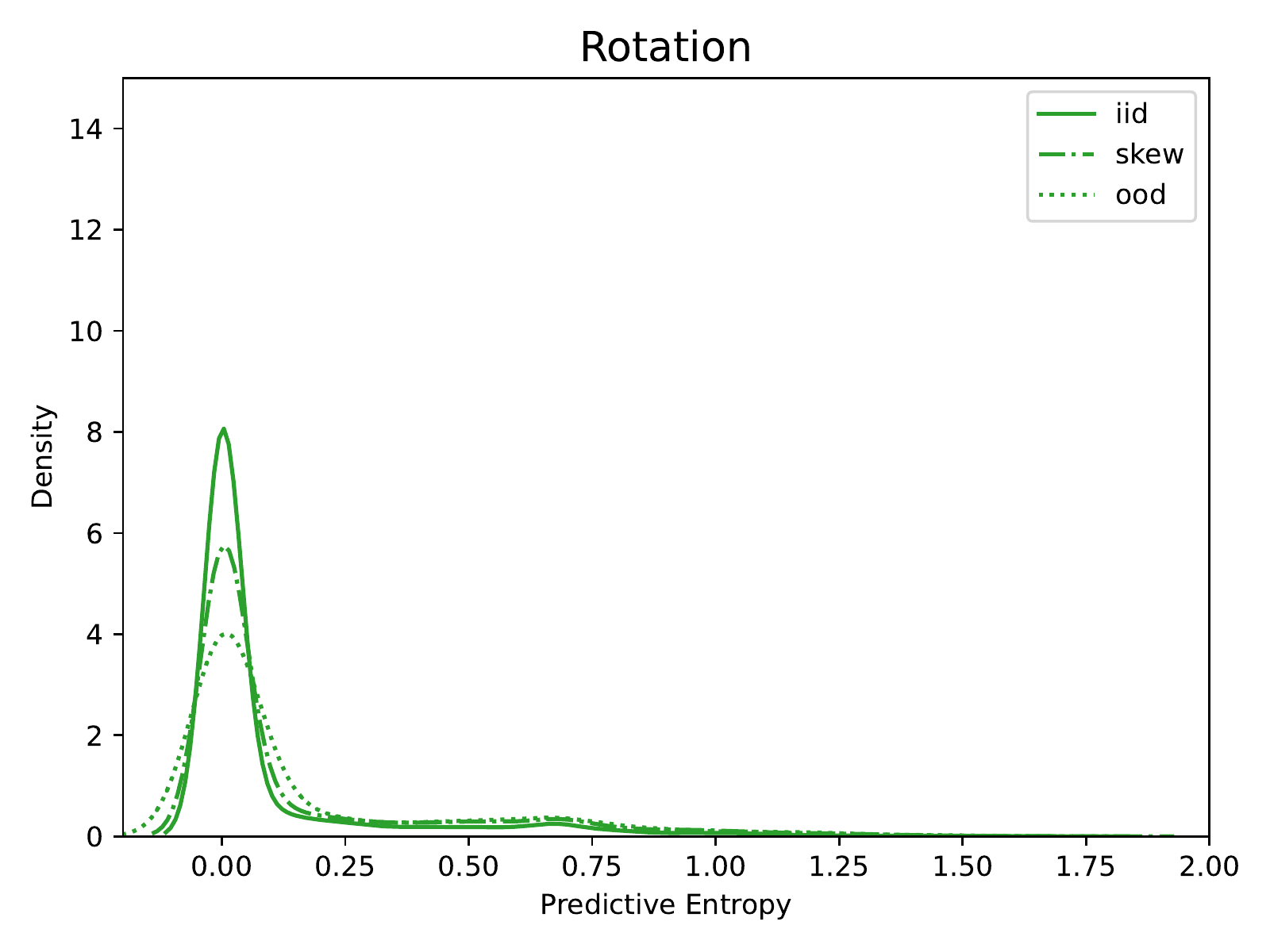}
}
\hspace{-0.15in}
\subfloat{
  \includegraphics[width=0.208\linewidth]{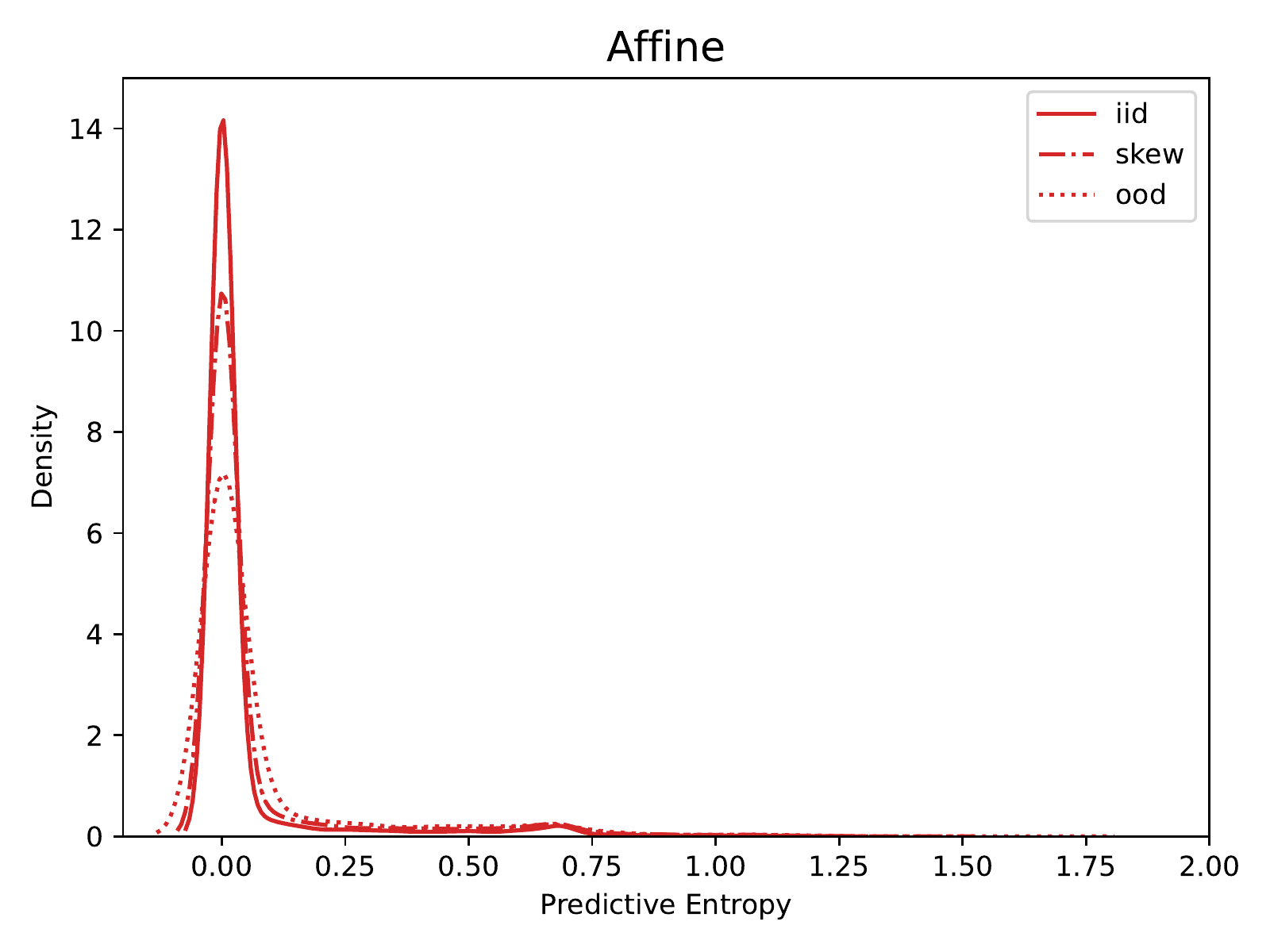}
}
\hspace{-0.15in}
\subfloat{
  \includegraphics[width=0.208\linewidth]{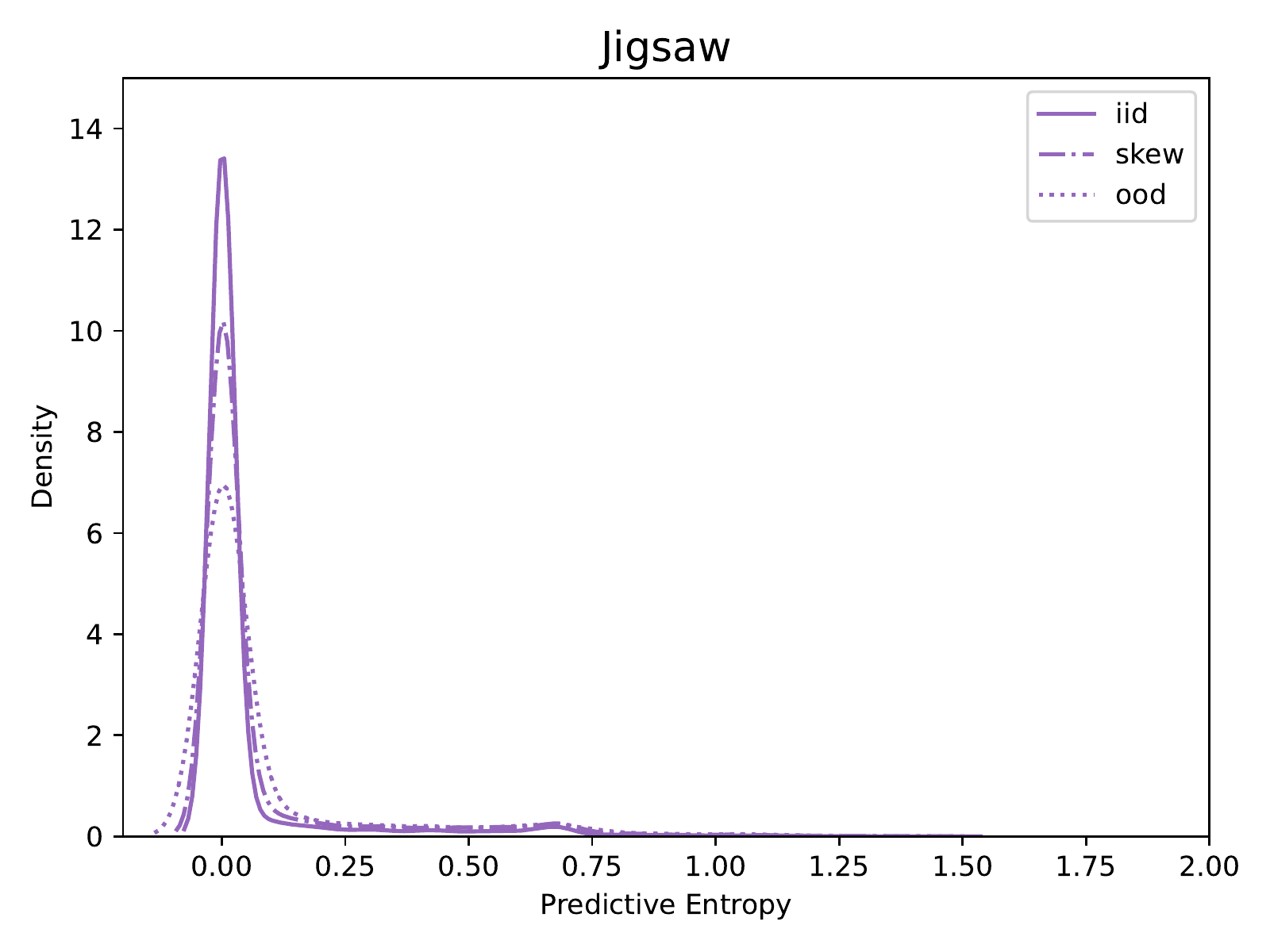}
}
\caption{Histograms of predictive entropy for Wider Resnet-28-10 architecture trained on CIFAR10 dataset and test on in of distribution (solid lines), out of distribution with CIFAR-10 Skew by "brightness" type (dotted lines), and the real-world out of distribution with CIFAR-10.1~v6 (dashed lines) (Zoom in for details).}
\label{fig:dPDF-diagram}
\vspace{-0.2in}
\end{figure}

\textbf{The bad uncertainty estimate of all \acrshort{SSL} methods.} In Section~\ref{sec:experiments}, we have observed all methods can not provide informative information about distributional shifts since their predictive entropy is often close to zero in the out-of-distribution test set. Figure~\ref{fig:dPDF-diagram} illustrates in more detail this bad uncertainty estimate for each method by comparing the histogram of their predictive entropy for in-of-distribution (solid lines), artificial out-of-distribution with the skew is "brightness" type in CIFAR-10 (dotted lines), and real-world out-of-distribution in CIFAR-10.1 v6 (dashed lines). Ideally, a reliable model should only provide predictive entropy close to zero in in-of-distribution data, and high entropy for out-of-distribution data. However, histograms for all methods only show the density value is lower while their entropy value does not increase too much under the out-of-distributional shifts, confirming the hypothesis that these methods do not good enough for a good reliability model in Section~\ref{sec:experiments}.%